\documentclass[journal]{IEEEtran}

\usepackage{cite}
\usepackage{amsmath,amssymb,amsfonts}
\usepackage{algpseudocode}
\usepackage{graphicx}
\usepackage{hyperref}
\hypersetup{hidelinks=true}
\usepackage[linesnumbered,ruled,vlined]{algorithm2e}
\usepackage{textcomp}
\usepackage{xcolor}
\usepackage{mathptmx}
\usepackage{multirow}
\usepackage{booktabs}
\usepackage[justification=centering]{caption}
\usepackage{makecell}
\usepackage{multicol}
\usepackage{setspace}
\usepackage{lipsum} 
\usepackage{subfigure}
\usepackage{tabularx}
\usepackage{float}
\usepackage{bm}
\usepackage{etoolbox}
\begin{document}

\title{Fuzzy Information Evolution with Three-Way Decision in Social Network Group Decision-Making}

\author{Qianlei Jia, Xinliang Zhou, Ondrej Krejcar, Enrique Herrera-Viedma (Fellow, IEEE)
\thanks{Corresponding author: Qianlei Jia}
\thanks{Qianlei Jia is with the Department of Electrical Engineering, Chalmers University of Technology, 41258, Gothenburg, Sweden and the School of Automation, Northwestern Polytechnical University, 710072, China, (e-mail: jiaql@mail.nwpu.edu.cn).}
\thanks{Xinliang Zhou is with the College of Computing and Data Science, Nanyang Technological Univeristy, 639798, Singapore, (e-mail: xinliang001@e.ntu.edu.sg).}
\thanks{Ondrej Krejcar is with the Faculty of Informatics and Management, University of Hradec Kralove, 500 03 Hradec Kralove, Czech Republic (e-mail: ondrej.krejcar@uhk.cz).}
\thanks{Enrique Herrera-Viedma is with the Andalusian Research Institute in Data
Science and Computational Intelligence and the Department of Computer
Science and Artificial Intelligence, University of Granada, 18071 Granada,
Spain, (e-mail: viedma@decsai.ugr.es)}
\thanks{This work was supported by the National Natural Science Foundation
of China (No. 62073266), by the grant PID2022-139297OB-I00 funded by 
MICIU/AEI/10.13039/501100011033, ERDF/EU, Grant C-ING-165-UGR23 funded by Consejería de Universidad,
Investigación e Innovación and by ERDF Andalusia Program 2021-2027.}
}

\maketitle

\begin{abstract}
\par In group decision-making (GDM) scenarios, uncertainty, dynamic social structures, and vague information present challenges for traditional opinion dynamics models. To address these issues, this study proposes a novel social network group decision-making (SNGDM) framework that integrates three-way decision (3WD) theory, dynamic network reconstruction, and linguistic opinion representation. First, the 3WD mechanism is introduced to explicitly model hesitation and ambiguity in agent judgments, thereby preventing irrational decisions. Second, a connection adjustment rule based on opinion similarity is developed, enabling agents to adaptively update their communication links and better reflect the evolving nature of social relationships. Third, linguistic terms are used to describe agent opinions, allowing the model to handle vague and incomplete information more effectively. Finally, an integrated multi-agent decision-making framework is constructed, which simultaneously considers individual uncertainty, opinion evolution, and network dynamics. The proposed model is applied to a multi-UAV cooperative decision-making scenario, where simulation results and consensus analysis demonstrate its effectiveness. Experimental comparisons further verify the \textcolor{black}{algorithm's advantages} in enhancing system stability and representing realistic decision-making behaviors.

\end{abstract}

\begin{IEEEkeywords}
Social network group decision-making (SNGDM), Three-way decision (3WD), Consensus analysis.
\end{IEEEkeywords}

\IEEEpeerreviewmaketitle
\renewcommand\arraystretch{1.2}

\section{Introduction}
\IEEEPARstart{I}{n} multi-agent decision-making, information is often highly uncertain. Traditional centralized information fusion and decision-making systems often face problems such as isolated information and poor coordination. To \textcolor{black}{enhance} local assessment capabilities and overall \textcolor{black}{situational} awareness, distributed information fusion and decision-making mechanisms based on multi-agent systems have \textcolor{black}{garnered} increasing attention in recent years \cite{zhu2025dynamic}.

\par Opinion dynamics models have been widely used as a fundamental framework for distributed information fusion in multi-agent systems. By simulating how individual agents update their information, these models effectively show collective behavior and the process of consensus formation in \textcolor{black}{an uncertain} environment \cite{dong2020consensus}. In \cite{shen2025hybrid}, the DeGroot and Hegselmann–Krause (HK) models were combined to propose a hybrid opinion dynamics (HOD) model in the large-scale group decision-making (LSGDM) framework. In \cite{chen2025competitive}, a self-confidence evolution model, which encompassed the self-confidence levels of one's group mates and the passage of time, was proposed. To consider the trust relationship in social network group decision making (SNGDM), a new discount-based uninorm trust propagation operator with 2-tuple linguistic was presented in \cite{xing2024trust}. Following the idea of the personalized individual semantics (PIS) model, the PIS-based linguistic opinions dynamics (PIS-LOD) model was proposed within the framework of bounded confidence effects \cite{liang2020linguistic}. In \cite{liang2016fusion}, a novel opinion dynamics model called interval opinion dynamics with dynamic bounded confidence was proposed, in which agents expressed their opinions in numerical intervals. By taking different numerical interval opinions and uncertainty tolerances into account, a numerical interval opinion dynamics model was proposed to investigate the process of forming collective opinions in a group of interacting agents \cite{dong2019numerical}. Inspired by the DeGroot model, a novel group decision-making (GDM) model with \textcolor{black}{opinion} evolution was established, and a consensus model with minimum adjustments was provided to obtain the optimal adjusted initial opinions and collective consensus opinion \cite{chen2019managing}. Besides, an opinion similarity mixed (OSM) model and a structural similarity mixed (SSM) model were proposed in \cite{wu2022mixed}, in which the strong and weak relations between individuals were considered. Based on the social network DeGroot model, the authors proved that the speed of consensus reaching was subject to the \textcolor{black}{most significant} self-confidence level of opinion followers and the speed of consensus reaching was subject to the top two self-confidence levels of opinion leaders \cite{ding2022consensus}. In \cite{pei2024conflict}, a conflict resolution method considering the trust propagation, conflict detection, and \textcolor{black}{alternative} selection was proposed.

\par As shown in the above analysis, opinion dynamics models have been widely applied across various fields. However, most of these models assume that agents must always make definite choices. For example, in the HK model, if the opinion difference between agents exceeds a given threshold, they will not interact or update their information. However, in many cases, the actual opinion difference may be only slightly larger than the threshold, yet it completely blocks information exchange, which is often unrealistic in practical situations. To overcome this limitation, this study incorporates the three-way decision (3WD) theory. 3WD has demonstrated unique advantages in various decision-making environments. In \cite{luo2019updating}, the update problem of 3WD with dynamic variation of scales in incomplete multi-scale information systems was studied, where the updating mechanisms of decision granules induced by the similarity relation were exploited through cut refinement and coarsening via attribute value taxonomies. Considering that the existing 3WD models could not effectively handle incomplete multiattribute decision-making (MADM) problems in real life, the preference of decision-makers for each alternative was considered, and the concept of predecisions was introduced, whereby an incomplete fuzzy decision system (IFDS) was obtained in \cite{zhan2021novel}. In \cite{wang2021three}, a kind of 3WD-MADM with probabilistic dominance relations was proposed, where the two state sets were developed by virtue of probabilistic dominance classes. Also, a probabilistic linguistic 3WD method based on regret theory (RT), namely PL-TWDR, was proposed for MADM problems with probabilistic linguistic term sets (PLTSs), considering the \textcolor{black}{practical} rationality of a decision-maker in complex decision environments \cite{zhu2023probabilistic}. In \cite{lang2019three}, the concepts of positive, neutral, and negative alliances with two thresholds were first provided. \textcolor{black}{Then}, a three-way conflict analysis based on the Bayesian minimum risk theory was conducted, where examples were explored to show how the positive, neutral, and negative alliances were computed with a Pythagorean fuzzy loss function given by an expert.

\par Moreover, the structure of \textcolor{black}{the social network} is also essential in GDM. In \cite{gai2025transformation}, a cooperation index was formulated to identify the non-cooperative behavior, based on which a non-cooperative behavior transformation method was investigated, and a bidirectional feedback mechanism was provided for SNGDM. In \cite{li2022trust}, trust risks in the conflict-eliminating process (CEP) of SNGDM were addressed through third-party monitoring, and a trust risk analysis-based conflict-eliminating model for SNGDM was developed. In \cite{liu2024game}, an interaction indicator was exploited to represent peer interaction effort (PIE), individual social cooperation networks (ISCNs) were constructed with the log-sigmoid transition technique, and a novel game-theoretic expert importance evaluation model guided by cooperation effects was finally proposed. To address the challenge that decision-makers in a community setting \textcolor{black}{typically exhibit} complex social preferences and intricate social interactions, \textcolor{black}{a minimum cost consensus-based SNGDM approach was designed, considering altruism-fairness preferences and ordered trust propagation \cite{feng2024minimum}.}

\par Despite the wide use of opinion dynamics models in distributed decision-making, several important challenges remain. First, some classical models, such as the DeGroot and HK frameworks, assume that agents must always make a clear decision, usually to accept or reject \textcolor{black}{a particular opinion}, and do not provide a mechanism \textcolor{black}{to express hesitation or uncertainty explicitly}. This forces agents to take a position even when the available information is not enough, which may lead to premature consensus or extreme polarization. For example, in a public policy vote, if a person does not have enough information about the pros and cons of a proposal, they may prefer to stay undecided until more evidence is available. However, in many classical models, this person is still forced to move toward “support” or “oppose” in each update step. The HK model partly addresses this problem by introducing a confidence threshold to limit interactions, but its results are \textcolor{black}{susceptible} to small changes in this threshold. For example, when the threshold is set to 0.3, if the opinion difference between two agents is 0.29, they will interact; but if the difference is 0.31, they will refuse to interact. In reality, these two situations are almost the same, but the model produces completely different results. This “boundary effect” \textcolor{black}{renders the model} highly sensitive to threshold settings, which reduces its robustness in dynamic environments. Second, some studies treat the network structure as fixed or oversimplified. In reality, communication and influence networks change over time. \textcolor{black}{Agents interact more frequently with some individuals than others, and these connections evolve in response to factors such as opinion similarity, trust, or communication costs}. For example, on social media platforms, people may increase interaction if their opinions become closer, or reduce \textcolor{black}{contact and} even unfollow, if their views diverge significantly. \textcolor{black}{Models that maintain a static network cannot capture this adaptive connection change, which limits their ability to describe fundamental social interactions and coordination.}

Motivated by the challenges identified above, this paper proposes a novel information updating model. The main contributions of this study are as follows:

\begin{enumerate}
\item \textcolor{black}{Three-way decision (3WD) mechanism for hesitation and uncertainty. We design an opinion updating rule based on the three-way decision framework, where each agent can choose “accept,” “reject,” or “defer” in response to incoming information. This differs from classical models such as DeGroot or HK that force agents into binary decisions. The mechanism explicitly preserves hesitation intervals, reducing the risk of premature consensus and improving decision robustness under conflicting or incomplete information.}
\item \textcolor{black}{Dynamic social network reconstruction driven by opinion similarity. We introduce a probabilistic link creation and deletion rule that updates the social network structure according to opinion distances between agents. This adaptive adjustment enables agents to strengthen ties with similar peers and weaken ties with divergent ones, capturing the evolving nature of real-world communication patterns.}
\item \textcolor{black}{Linguistic opinion representation for vague and subjective information. We represent opinions using linguistic terms mapped from numerical values, instead of using only numerical states. This makes the model better suited for situations where information is subjective, vague, or incomplete. The approach improves interpretability and aligns simulation outputs more closely with real-world decision-making contexts, as validated in multi-agent test scenarios.}
\end{enumerate}

\par The paper's structure is as follows: In Section \uppercase\expandafter{\romannumeral2}, the relevant concepts are introduced. In Section \uppercase\expandafter{\romannumeral3}, the new model is proposed. In Section \uppercase\expandafter{\romannumeral4}, a series of examples are presented. In Section \uppercase\expandafter{\romannumeral5}, the comparison is conducted. The conclusion is given in Section \uppercase\expandafter{\romannumeral6}.
\section{Preliminaries}
\subsection{Linguistic Term Set}
\par \emph{Definition 1:} Let $H=\lbrace h_i \vert i=0,1,\cdots,2\Phi,\vert \Phi\in N^* \rbrace$ be finite and completely ordered linguistic term sets (LTSs). $h_i$ represents a possible value for a linguistic variable. $h_i$ and $h_j$ should satisfy \cite{9468885, 1974The}: 
\par (1) $h_i \leq h_j$ if and only if $i \leq j$; 
\par (2) The negation operation $neg(h_i)=h_j$ if $i + j=2\Phi$;
\par (3) If $i \geq j$, then $\max \lbrace h_i,h_j \rbrace=h_i$;
\par (4) If $i \geq j$, then $\min \lbrace h_i,h_j \rbrace=h_j$.

\subsection{Linguistic Scale Functions (LSFs)}
\par \emph{Definition 2 \cite{6798659}:} For the linguistic term $h_j$ in $H=\lbrace h_i \vert i=0,1,\cdots,2\Phi,\vert \Phi\in N^* \rbrace$, $h_j$ represents a possible value for a linguistic variable. If $\theta\in [0,\,1]$ is a numerical value, then the function $f:h_j \rightarrow \theta_j$, which conducts the mapping from $h_j$ to $\theta_j\;(j=0,\;1,\;\cdots,\;2t)$, is defined as follows:
\par \begin{equation} 
\theta_j=\left\{ \begin{array}{rcl} \frac{a^t-a^{t-j}}{2a^t-2}, & & {0 \leq j \leq t}\\ 
\frac{a^t+a^{j-t}-2}{2a^t-2}, & & {t<j\leq2t } \end{array} 
\right. \end{equation}

\subsection{DeGroot Model}
\par \emph{Definition 3 \cite{9788051}:} In the DeGroot model, the agents unconditionally trust \textcolor{black}{each other}. Suppose that $E=\lbrace e_1,e_2,\cdots,e_N \rbrace$ indicates the agents. $t\in \lbrace 1,2,\cdots,T \rbrace$ indicates the discrete variable representing the number of rounds of opinion evolution. For $i\in N$, $y_i(t)\in [0,1]$ is the numeric opinion at time $t$. The weight offered by the agent $e_i$ to $e_j$ is expressed as $\omega_{j}$, where $0\leq \omega_{j}\leq 1$ and $\sum_{j=1}^N \omega_{j}=1$ for all $i\in N$. The \textcolor{black}{opinion} evolution rule is:
\begin{equation}
y_i(t+1)=\sum_{j=1}^N \omega_{j}y_j(t).
\end{equation}
\subsection{HK Bounded Confidence Model}
\par \emph{Definition 4 \cite{LI2021115479, 9107399}:} Suppose that $E=\lbrace e_1,e_2,\cdots,e_k \rbrace$ indicates the agents. For $e_i\in E$, $x_i(t)\in [0,1]$ is the numeric opinion at time $t$. $X=\lbrace x_1(t),x_2(t),\cdots,x_k(t) \rbrace$ is the opinions of agents express at time $t$. $\varepsilon$ indicates the confidence threshold value. If all
agents hold the same confidence threshold $\varepsilon$, the HK model is homogeneous; otherwise, it is heterogeneous.
\par \emph{Step 1:} Determine the confidence set.
\par Let $I(e_i,X(t))$ be the confidence set of $e_i$ at time $t$.
\begin{equation}
I(e_i,X(t))=\lbrace e_j\vert \vert x_i(t)-x_j(t)\vert\leq \varepsilon,e_j\in E\rbrace,\;\; i=1,2,\cdots,k
\end{equation}
\par \emph{Step 2:} Calculate the weights.
\par When the agent $e_i$ \textcolor{black}{updates the opinion} at time $t$, the weight of the agent $e_j$ is:
\begin{equation} 
w_{ij}(t)=
\left\{
\begin{array}{c@{\quad}l}
\frac{1}{\# I(e_i,Y(t))}, & \text{if } e_j \in I(e_i,Y(t)) \\
0, & \text{otherwise}
\end{array}
\right.
\end{equation}
\par where $\# I(e_i,Y(t))$ is the cardinality of $I(e_i,X(t))$.
\par \emph{Step 3:} Opinion evolution.
\par Let $x_i(t+1)$ \textcolor{black}{indicate the opinion} of $e_i$ at time $t+1$, then
\begin{equation}
x_i(t+1)=w_{i1}x_1(t)+w_{i2}x_2(t)+\cdots+w_{ik}x_k(t)
\end{equation}
\subsection{Social Network}
\par \emph{Definition 5 \cite{2017ManagDonging, 2014Anetwork, 9057368Behavior}:} A social network is $G(V,E)$. $V=\lbrace v_t,\;t \in M \rbrace$ expresses vertices. $E=\lbrace e_{ts},\;t,s \in M \rbrace$ indicates the edges. The edge $e_{ts}=(v_t,v_s)$ indicates the relationship between $v_t$ and $v_s$. $F=(f_{ts})_{m \times m}$ describes the social network. 
\begin{equation} 
f_{ts}=\left\{ \begin{array}{rcl} 1 , & & {(v_t,v_s)\in E}\\ 
0, & & {otherwise } \end{array} 
\right. \end{equation}
\par \emph{Definition 6 \cite{1995SocialContemporary, DING2019251}:} For $G(V,E)$, the following concepts are introduced.
\par Density of social network: \par It is the ratio of the number of direct pairwise relationships to the possible number of direct relationships.
\par In-degree $C_{id}(v_t)$ of the vertex $v_t$: 
\par It is employed to describe the reputation of a node, which reflects the integration force.
\par Out-degree $C_{od}(v_t)$ of the vertex $v_t$: 
\par It is used to describe the communicative nature of a node, which reflects the radiation.
\par The final centrality degree is $\gamma=\frac{C_{id}(v_t)+C_{od}(v_t)}{2(m-1)}$.
\subsection{3WD Model}
\par \emph{Definition 7 \cite{wang2022three}:} The 3WD model includes two states $X = \{C, \neg C\}$ and three actions $H = \{h_A, h_D, h_R\}$, where $C$ and $\neg C$ represent for a ``good" state and a ``bad" state, respectively. $h_A$, $h_D$, and $h_R$ represent the three domains in classifying an alternative $a_i$, that is, $a_i \in POS(C), \, a_i \in BOU(C), \, \text{and} \; a_i \in NEG(C)$, respectively. Furthermore, $POS(C), \, BOU(C), \, \text{and} \; NEG(C)$ correspond to the positive domain, the boundary domain\textcolor{black}{, and} the negative domain, respectively. When $a_i \in C$, let $\lambda_{AP}$, $\lambda_{DP}$, and $\lambda_{RP}$ denote the losses of taking the actions $h_A$, $h_D$, and $h_R$. Similarly, when $a_i \in \neg C$, let $\lambda_{AN}$, $\lambda_{DN}$, and $\lambda_{RN}$ represent the losses of taking the actions $h_A$, $h_D$, and $h_R$. Then, the loss functions of $a_i$ are given in Table \uppercase\expandafter{\romannumeral1}, where $C(P)$ and $\neg C(P)$ are represented by $P$ and $N$ standing for $C$ and $\neg C$, respectively. For any alternative $a_i \in A$, the equivalence class of $a_i$ is represented as $[a_i]$. $Pr(C \mid [a_i])$ and $Pr(\neg C \mid [a_i])$ represent the conditional probability of the alternative $a_i$ belongs to and not belonging to $C$, respectively. Then, the expected losses $R(h_I \mid [a_i])$ for the three actions ($I = A; D; R$) are:

\begin{equation}\nonumber
\begin{aligned}
R(h_A \mid [a_i])=\lambda_{AP} Pr(C \mid [a_i])+\lambda_{AN}Pr(\neg C \mid [a_i]);\\
R(h_D \mid [a_i])=\lambda_{DP} Pr(C \mid [a_i])+\lambda_{DN}Pr(\neg C \mid [a_i]);\\
R(h_R \mid [a_i])=\lambda_{RP} Pr(C \mid [a_i])+\lambda_{RN}Pr(\neg C \mid [a_i]).
\end{aligned}
\end{equation}
\begin{table}[!h]
\centering
\caption{Loss Functions of $a_i$.}
\setlength{\tabcolsep}{7.5mm}
\renewcommand\arraystretch{1.1}
\begin{tabular}{ccc}
\bottomrule
 &$C(P)$&$\neg C(N)$\\
\hline
$h_A$& $\lambda_{AP}$ & $\lambda_{AN}$ \\
$h_D$& $\lambda_{DP}$ & $\lambda_{DN}$ \\
$h_R$& $\lambda_{RP}$ & $\lambda_{RN}$\\
\toprule
\end{tabular}
 \end{table} 
\par Afterwards, we can obtain the following rules by using the Bayesian theory of minimum losses.
\begin{equation}\nonumber
\begin{aligned}
(P)\; \text{If}\, R(h_A \mid [a_i])\leq R(h_D \mid [a_i])\, \text{and}\; R(h_A \mid [a_i])\leq R(h_R \mid [a_i]),\, \\ \text{decide}\; a_i \in POS(C);\quad\quad\quad\quad\quad\quad\quad\quad\quad\quad\quad\quad\quad\quad\quad\\
(B)\; \text{If}\, R(h_D \mid [a_i])\leq R(h_A \mid [a_i])\, \text{and}\; R(h_D \mid [a_i])\leq R(h_R \mid [a_i]),\, \\ \text{decide}\; a_i \in BOU(C);\quad\quad\quad\quad\quad\quad\quad\quad\quad\quad\quad\quad\quad\quad\quad\\
(N)\; \text{If}\, R(h_R \mid [a_i])\leq R(h_A \mid [a_i])\, \text{and}\; R(h_R \mid [a_i])\leq R(h_D \mid [a_i]),\, \\ \text{decide}\; a_i \in NEG(C).\quad\quad\quad\quad\quad\quad\quad\quad\quad\quad\quad\quad\quad\quad\quad
\end{aligned}
\end{equation}

\section{Linguistic 3WD Opinion Dynamics Model over \\Dynamic Social Networks}
\par To better model the interaction between opinion evolution and network structure in GDM, this paper proposes a co-evolutionary model that combines the 3WD mechanism with a dynamic network update process. Traditional models often use a binary decision rule, where agents either fully accept or completely reject neighbors based on a fixed confidence bound. However, such a hard-boundary approach cannot fully reflect the diversity of real-world social behavior. In this work, we introduce the 3WD mechanism, which divides the opinion distance between two agents into three regions: if the distance is less than a strict acceptance threshold $\alpha$, the neighbor is deterministically accepted; if the distance exceeds a rejection threshold $\beta$, the neighbor is rejected; and if the distance falls between $\alpha$ and $\beta$, the agent accepts the neighbor with a certain probability. This probability is controlled by an exponential decay function $p = \exp(-\lambda(d_{ij} - \alpha))$, where $d_{ij}$ is the opinion distance and $\lambda$ is a decay factor. This design reflects the psychological tendency that people are more likely to accept opinions that are closer to their own, while opinions farther away are less likely to be accepted. The 3WD mechanism allows the model to capture hesitation, selective trust, and partial agreement, which are common in \textcolor{black}{fundamental social interactions}. In addition, to break the limitation of a static network, we introduce a dynamic update mechanism based on cognitive similarity. This mechanism allows the connections between agents to change over time, depending on their current opinions. Specifically, if two agents are not connected but their opinion distance is smaller than a threshold $\delta_{\text{add}}$, a link is added between them with probability $p_{\text{add}}$. On the other hand, if two connected agents show a \textcolor{black}{significant opinion difference} that exceeds $\delta_{\text{cut}}$, their link may be removed with probability $p_{\text{cut}}$. In this way, the network structure evolves according to the cognitive relationships among agents.

\par We consider a group of $N$ agents whose opinions are expressed using a finite linguistic term set $H = \{ h_0, h_1, \dots, h_{2\Phi} \}$. Each term $h_j \in H$ is mapped to a real-valued scalar in the interval $[0,1]$ via a predefined nonlinear function $f: H \rightarrow [0,1]$. Let $x_i(t) \in H$ denote the linguistic opinion of agent $i$ at time $t$, and let $\theta_i(t) = f(x_i(t))$ be the corresponding numerical representation.
\par \textbf{Notation:}
\begin{itemize}
    \item $H = \{h_0, h_1, \dots, h_{2\Phi}\}$: Linguistic term set
    \item $f(h_j) = \theta_j$: Mapping from $h_j$ to $[0,1]$
    \item $x_i(t) \in H$: Linguistic opinion of agent $i$ at time $t$
    \item $\theta_i(t) = f(x_i(t))$: Numerical opinion
    \item $A_{ij}(t)$: Social adjacency matrix ($1$ if connected, $0$ otherwise)
    \item $\alpha$, $\beta$: Three-way decision thresholds
    \item $\lambda$: Probability decay parameter
    \item $\delta_{\text{add}}, \delta_{\text{cut}}$: Network update distance thresholds
    \item $p_{\text{add}}, p_{\text{cut}}$: Rewiring probabilities
\end{itemize}

\paragraph{Initialization}
Each agent is randomly assigned an initial linguistic opinion $x_i(0) \in H$, which is then mapped to its corresponding numeric value $\theta_i(0) = f(x_i(0))$. The initial social network is represented by an adjacency matrix $A(0)$, where $A_{ij}(0) = 1$ if agents $i$ and $j$ are connected, and $A_{ij}(0) = 0$ otherwise.

\begin{algorithm}[htbp]
\caption{Linguistic 3WD Opinion Dynamics over Dynamic Social Network}
\KwIn{\newline
Linguistic term set $H$, mapping function $f$, initial network $A(0)$, parameters $\alpha, \beta, \lambda, \delta_{\text{add}}, \delta_{\text{cut}}, p_{\text{add}}, p_{\text{cut}}, T_{\max}, \varepsilon$}
\KwOut{Final linguistic opinions $x_i$ for all agents}

\textbf{Initialization:} \newline
\ForEach{agent $i \in \{1, \dots, N\}$}{
    Randomly assign $x_i(0) \in H$\;
    Compute $\theta_i(0) = f(x_i(0))$\;
}

\For{$t = 0$ \KwTo $T_{\max}$}{
    \ForEach{agent $i \in \{1, \dots, N\}$}{
        Initialize neighbor set $\mathcal{N}_i(t) \gets \emptyset$\;
        \ForEach{$j \ne i$ \textbf{and} $A_{ij}(t) = 1$}{
            Compute $d_{ij} = |\theta_i(t) - \theta_j(t)|$\;
            \eIf{$d_{ij} \leq \alpha$}{
                $\mathcal{N}_i(t) \gets \mathcal{N}_i(t) \cup \{j\}$ \tcp*[r]{Accept}
            }{\eIf{$d_{ij} \geq \beta$}{
                \textbf{continue} \tcp*[r]{Reject}
            }{
                Compute $p = \exp(-\lambda(d_{ij} - \alpha))$\;
                Sample $r \sim \mathcal{U}(0,1)$\;
                \If{$r < p$}{
                    $\mathcal{N}_i(t) \gets \mathcal{N}_i(t) \cup \{j\}$ \tcp*[r]{Accept with uncertainty}
                }
            }}
        }
        \eIf{$\mathcal{N}_i(t) \neq \emptyset$}{
            $\theta_i(t+1) = \frac{1}{|\mathcal{N}_i(t)|} \sum\limits_{j \in \mathcal{N}_i(t)} \theta_j(t)$\;
        }{
            $\theta_i(t+1) = \theta_i(t)$ \tcp*[r]{No update}
        }
        Map back to linguistic term:\;
        $x_i(t+1) = \arg\min\limits_{h_j \in H} |f(h_j) - \theta_i(t+1)|$\;
    }

    \tcp{Step 5: Dynamic Network Rewiring}
    \ForEach{pair $(i,j)$ with $i \ne j$}{
        Compute $d_{ij}(t) = |\theta_i(t) - \theta_j(t)|$\;
        \If{$A_{ij}(t) = 0$ \textbf{and} $d_{ij}(t) < \delta_{\text{add}}$}{
            Sample $r \sim \mathcal{U}(0,1)$\;
            \If{$r < p_{\text{add}}$}{
                Set $A_{ij}(t+1) = A_{ji}(t+1) = 1$ \tcp*[r]{Add link}
            }
        }
        \If{$A_{ij}(t) = 1$ \textbf{and} $d_{ij}(t) > \delta_{\text{cut}}$}{
            Sample $r \sim \mathcal{U}(0,1)$\;
            \If{$r < p_{\text{cut}}$}{
                Set $A_{ij}(t+1) = A_{ji}(t+1) = 0$ \tcp*[r]{Remove link}
            }
        }
    }

    Check convergence:\;
    \If{$\max\limits_i |\theta_i(t+1) - \theta_i(t)| < \varepsilon$}{
        \textbf{break}\;
    }
}
\end{algorithm}

\paragraph{Opinion Updating}
At each discrete time step $t = 1, 2, \dots, T_{\max}$, every agent $i$ updates its opinion based on the opinions of its neighbors. Specifically, agent $i$ evaluates each connected agent $j$ (i.e., $A_{ij}(t) = 1$) by computing the opinion distance $d_{ij}(t) = |\theta_i(t) - \theta_j(t)|$. Then, a 3WD strategy is applied:
\begin{itemize}
    \item If $d_{ij}(t) \leq \alpha$, agent $j$ is accepted as a trusted neighbor.
    \item If $d_{ij}(t) \geq \beta$, agent $j$ is rejected.
    \item If $\alpha < d_{ij}(t) < \beta$, agent $j$ is accepted with probability $p = \exp(-\lambda (d_{ij}(t) - \alpha))$.
\end{itemize}
Let $\mathcal{N}_i(t)$ denote the set of accepted neighbors of agent $i$ at time $t$. If $\mathcal{N}_i(t)$ is nonempty, agent $i$ updates its numerical opinion by averaging:
\[
\theta_i(t+1) = \frac{1}{|\mathcal{N}_i(t)|} \sum_{j \in \mathcal{N}_i(t)} \theta_j(t).
\]
If $\mathcal{N}_i(t)$ is empty, then $\theta_i(t+1) = \theta_i(t)$ and the opinion remains unchanged. After the numerical update, the agent's opinion is mapped back to the nearest linguistic term:
\[
x_i(t+1) = \arg\min_{h_j \in H} |f(h_j) - \theta_i(t+1)|.
\]

\paragraph{Dynamic Network Rewiring}
Following the opinion update, the social network evolves based on the opinion differences among agents. For each pair of agents $(i,j)$:
\begin{itemize}
    \item If $A_{ij}(t) = 0$ and $|\theta_i(t) - \theta_j(t)| < \delta_{\text{add}}$, a link is added with probability $p_{\text{add}}$.
    \item If $A_{ij}(t) = 1$ and $|\theta_i(t) - \theta_j(t)| > \delta_{\text{cut}}$, the existing link is removed with probability $p_{\text{cut}}$.
\end{itemize}
The adjacency matrix is then updated to $A(t+1)$ accordingly.

\paragraph{Termination Condition}
The iteration continues until the maximum opinion change between two successive steps is smaller than a convergence threshold $\varepsilon$, i.e.,
\[
\max_{i} |\theta_i(t+1) - \theta_i(t)| < \varepsilon,
\]
or until the maximum number of iterations $T_{\max}$ is reached.

\paragraph{Output}
The final linguistic opinion $x_i$ for each agent is obtained, representing the converged state or the state at termination.

\section{Example} 
\par In recent years, with the development of multi-UAV cooperative technology, UAV swarm intelligence systems have demonstrated a high level of autonomy and efficient information-sharing capabilities across various fields. It is worth noting that this study is inspired by the research presented in \cite{huang2025multiple}. As shown in Fig. 1, a group of \( n \) UAVs is cooperatively circumnavigating a group of \( m \) unknown targets on a horizontal plane. In this paper, we take post-earthquake rescue in mountainous areas as an example. Multiple UAVs are often deployed to search for survivors across large regions. In this setting, the model based on collaborative decision-making and opinion dynamics is essential. Each UAV acts as an intelligent \textcolor{black}{agent and shares} the linguistic information with the system. As these agents communicate, they update the evaluation information of the situation.
\begin{figure}[!htb]
\centering
\includegraphics[scale=0.5]{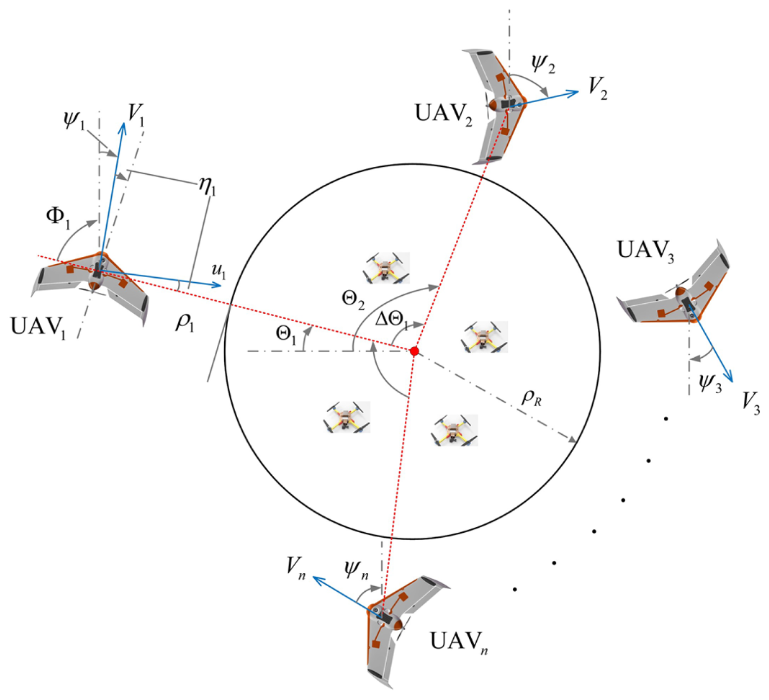}
\caption{Circumnavigation of $m$ Targets by $n$ UAVs (Red Center Represents the
Geometric Centroid of $m$ Targets) \cite{huang2025multiple}.}
\label{debris}
\end{figure}

\par To evaluate the proposed linguistic 3WD opinion dynamics model, we conduct simulations with a fixed number of agents and predefined model parameters. Specifically, the number of agents was set to $N = 20$. The linguistic term set consisted of $2\Phi + 1$ elements, with $\Phi = 3$, and the non-linear mapping parameter was set to $a = 2$. The 3WD is controlled by two thresholds: the strict acceptance threshold $\alpha = 0.3$, and the strict rejection threshold $\beta = 0.6$. Agents falling in the uncertain region $(\alpha, \beta)$ were accepted with a probability that decays exponentially with the distance from $\alpha$, governed by a decay factor $\lambda = 10$. The opinion inertia parameter was set to $\omega = 0$, meaning that agents fully adopt the weighted average of their accepted neighbors' opinions without retaining their previous value. For the dynamic network update process, we adopted an additive threshold $\delta_{\text{add}} = 0.15$ and a cut-off threshold $\delta_{\text{cut}} = 0.45$. Links were added or removed with probabilities $p_{\text{add}} = 0.5$ and $p_{\text{cut}} = 0.5$, respectively. The simulation ran for at most $T_{\max} = 10$ time steps, or until convergence was reached, defined by a maximum opinion change less than a tolerance of $\varepsilon = 10^{-3}$.

\par The initial linguistic opinions of the 20 agents at time step $t = 0$ are given by:
\[
\begin{aligned}
x_1(0) &= h_0, \quad &x_2(0) &= h_3, \quad &x_3(0) &= h_6, \quad &x_4(0) &= h_0, \\
x_5(0) &= h_0, \quad &x_6(0) &= h_0, \quad &x_7(0) &= h_5, \quad &x_8(0) &= h_3, \\
x_9(0) &= h_3, \quad &x_{10}(0) &= h_3, \quad &x_{11}(0) &= h_5, \quad &x_{12}(0) &= h_1, \\
x_{13}(0) &= h_0, \quad &x_{14}(0) &= h_5, \quad &x_{15}(0) &= h_3, \quad &x_{16}(0) &= h_0, \\
x_{17}(0) &= h_4, \quad &x_{18}(0) &= h_3, \quad &x_{19}(0) &= h_4, \quad &x_{20}(0) &= h_3.
\end{aligned}
\]

\begin{figure}[!htp]
\centering
\includegraphics[width=3.0in]{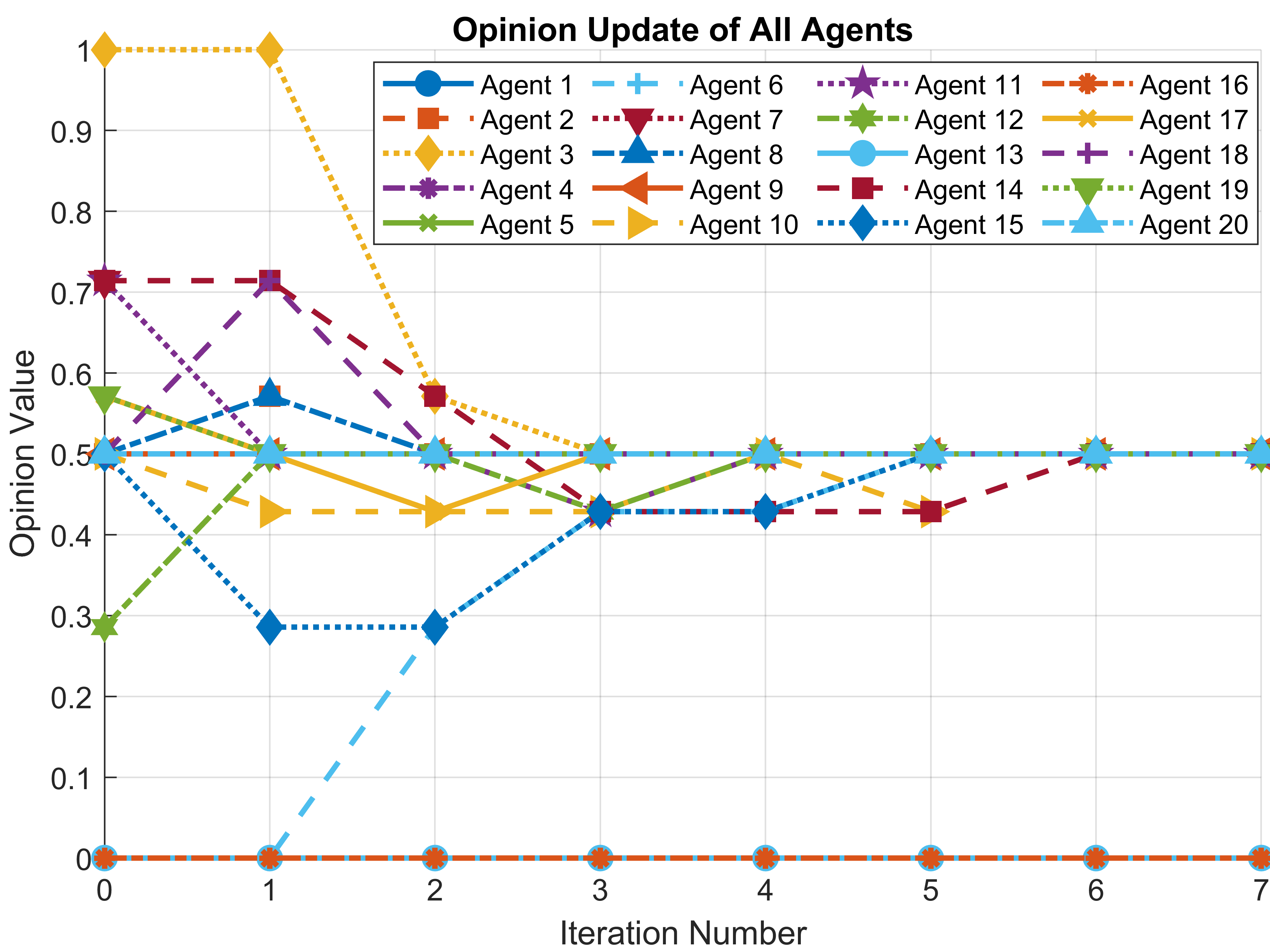}
\caption{Opinions of All Agents Across Iterations.}
\label{fig:opinion_trajectories}
\end{figure}
\begin{figure}[!htp]
\centering
\includegraphics[width=3.0in]{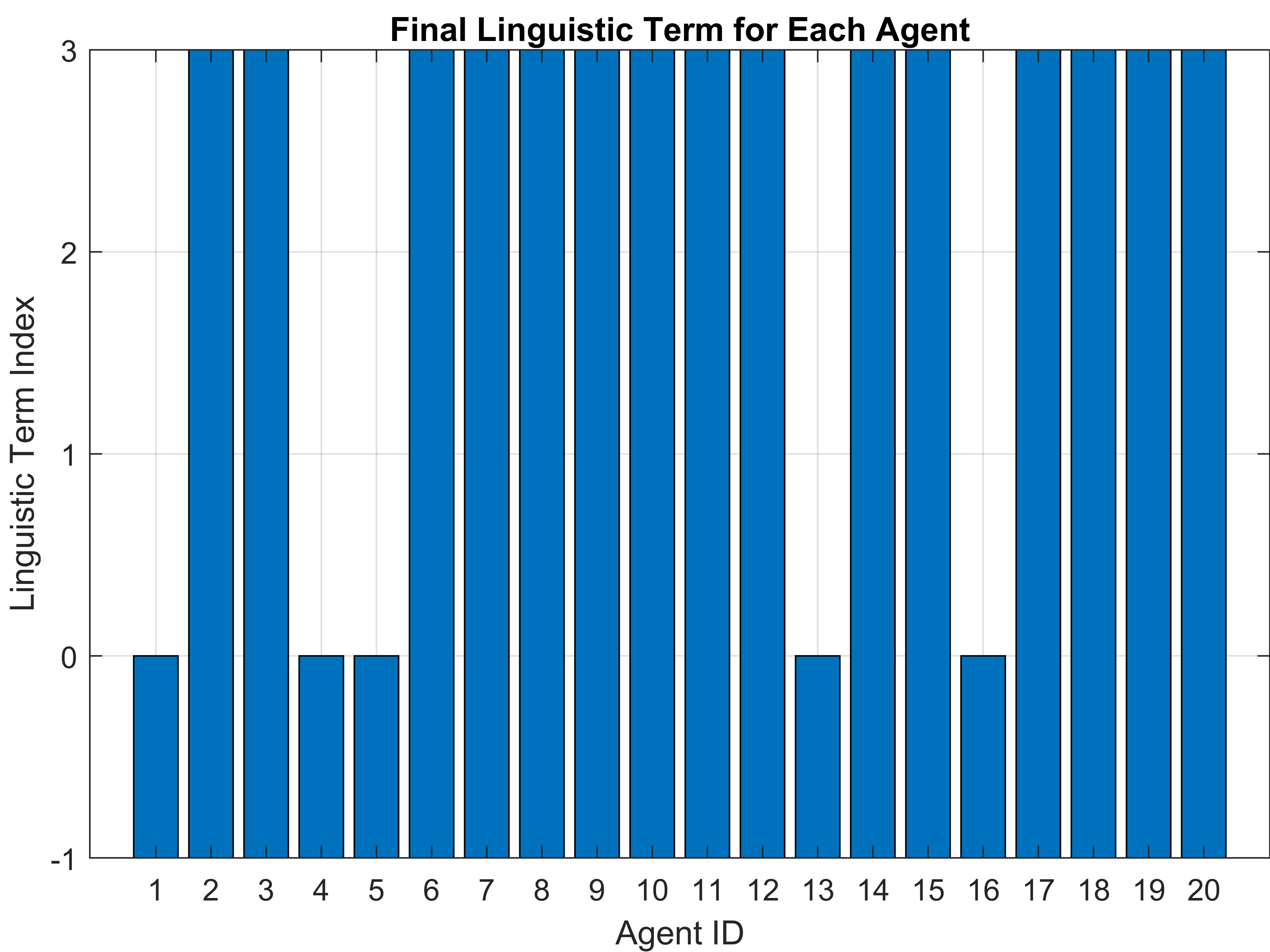}
\caption{Final Linguistic Term for Each Agent.}
\label{fig:final_results}
\end{figure}
\par
\begin{figure*}[!htp]
\centering
\includegraphics[width=7.0in]{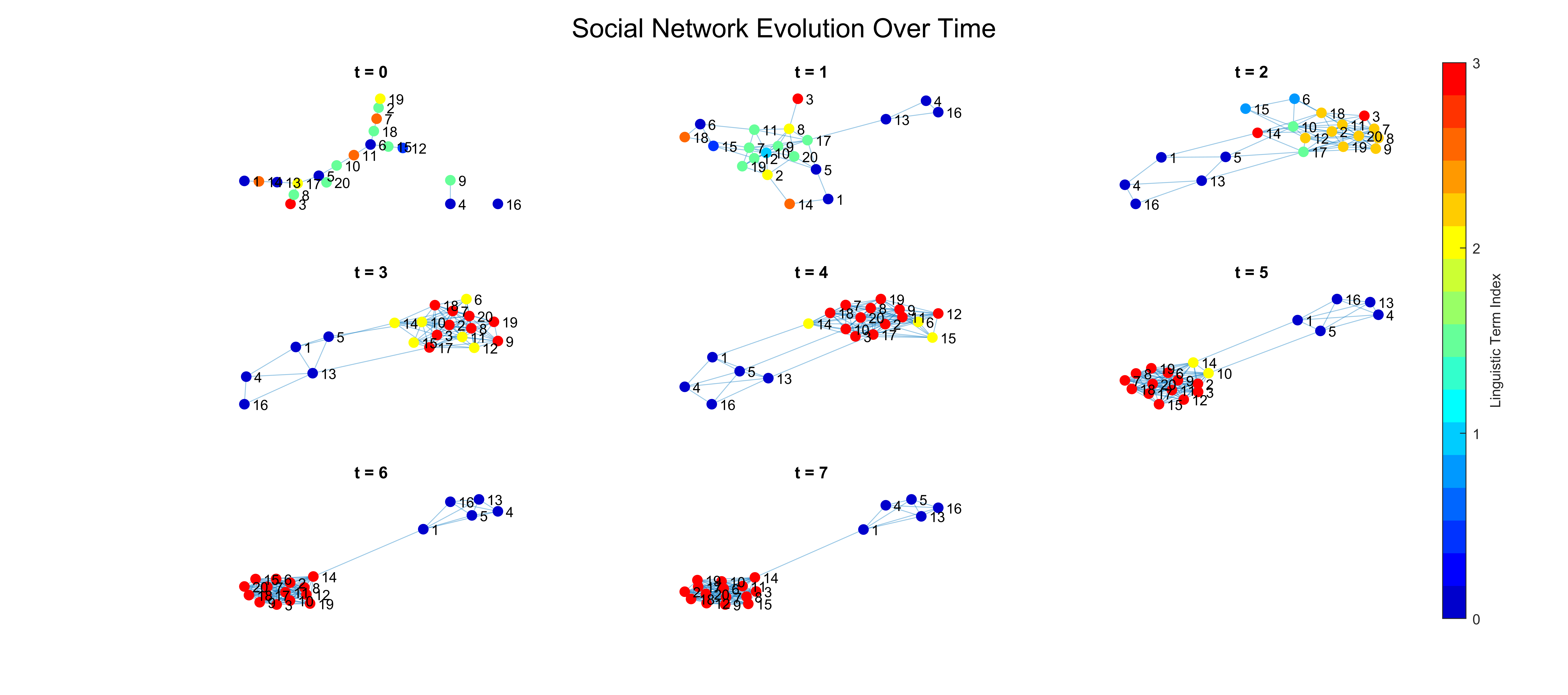}
\caption{Temporal Evolution of the Social Network and Linguistic Opinions.}
\label{fig:social_network}
\end{figure*}

\begin{figure}[!htp]
\centering
\includegraphics[width=3.0in]{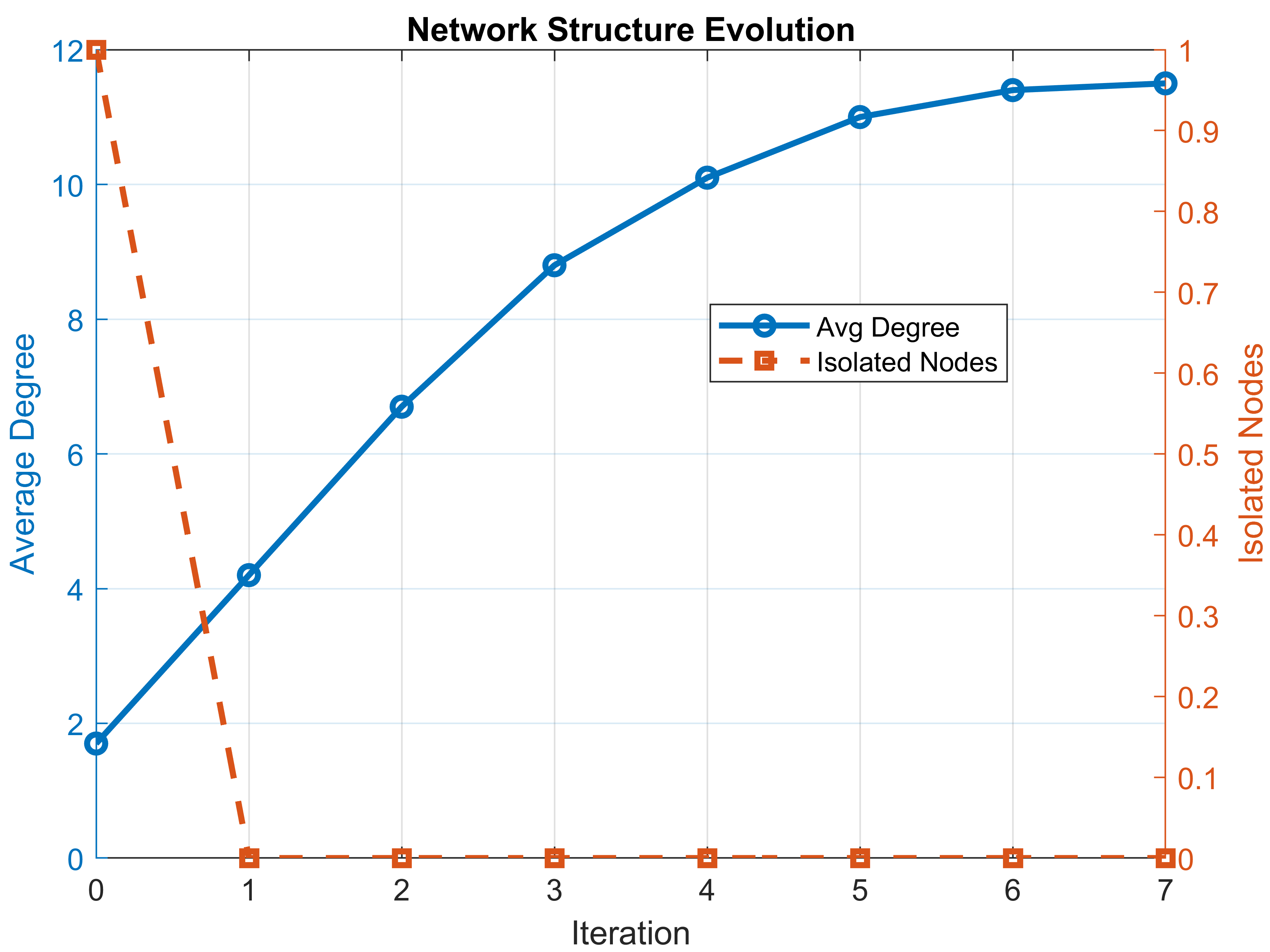}
\caption{Network Connectivity Trends During Evolution.}
\label{fig:network_evolution}
\end{figure}

 Fig.~\ref{fig:opinion_trajectories} shows how the opinions of all agents change over time. \textcolor{black}{We can see that several agents adjusted their opinions clearly in the early stages, which indicates active communication and belief change at the outset}. Most of these changes happened within the first four iterations. After that, the opinions became more stable. By the fifth iteration, most agents had nearly stopped changing. At the seventh iteration, all opinions stayed the same, which means the system reached a stable state and no more updates are needed. The final linguistic term for each agent is shown in Fig.~\ref{fig:final_results}. Fig.~\ref{fig:social_network} presents the evolution of the social network and agent opinions from iteration 0 to 7. In each subplot, nodes represent agents and edges represent social ties. Node colors indicate the linguistic term each agent holds at that time step. It can be seen that agents with similar opinions gradually cluster together as the simulation progresses. From iteration 5 onwards, the network evolves into two densely connected subgroups, each associated with an opinion. Fig.~\ref{fig:network_evolution} shows how the network structure changes over the iterations. The average degree increases quickly from about 1.9 to nearly 12, which means that agents are building more links with each other \textcolor{black}{as the iteration goes on}. This suggests that agents are becoming more similar in their opinions, so new connections are more likely under the network update rule. Meanwhile, the number of isolated nodes drops from 1 to 0 in the first two steps and stays at zero after that. This means that all agents are connected to the network early on, and the structure remains fully connected during the process. These results show that the network becomes more connected over time.

\par \textcolor{black}{To quantitatively evaluate the consensus degree of the system, this paper adopts three complementary indicators. First, the opinion variance $
\mathrm{Var}(t) = \frac{1}{n}\sum_{i=1}^n \bigl(x_i(t)-\bar{x}(t)\bigr)^2$ is used to reflect the overall dispersion of the group, where a smaller variance indicates stronger convergence. Second, the opinion range $R(t) = \max_i x_i(t) - \min_i x_i(t)$ is applied to measure the extreme difference within the group, and $R(t)=0$ means that all agents have reached complete consensus. Finally, we introduce the normalized consensus index $ C_{\mathrm{AAD}}(t) = 1 - \frac{1}{n}\sum_{i=1}^n \frac{|x_i(t)-\bar{x}(t)|}{d_{\max}}$ based on the average absolute deviation, and $d_{\max}$ is the maximum possible deviation in the opinion space. This index is normalized to the interval $[0,1]$, and values closer to 1 indicate a higher level of consensus. By combining these three measures, the consensus level of the system can be comprehensively evaluated from the perspectives of overall convergence, extreme differences, and normalized description. In this case, we can obtain $\mathrm{Var}(7)=0.0469$, $R(7)=0.5$, and $C_{\mathrm{AAD}}(7)=0.625$. These values show that the opinions still have some dispersion (since the variance 0.0469 is not close to 0), and the largest difference between individuals is 0.5, which means there is a clear gap. At the same time, the normalized consensus index is 0.625, which indicates that the system has not reached full agreement but rather a state of partial consensus.}


\section{Comparison, Robustness Analysis, and Complexity Analysis}

\subsection{Impact of Parameters and Mechanisms on Proposed Model}
\subsubsection{Impact of Broad Acceptance Thresholds on Consensus Formation}
\begin{figure}[!htp]
\centering
\includegraphics[width=3.0in]{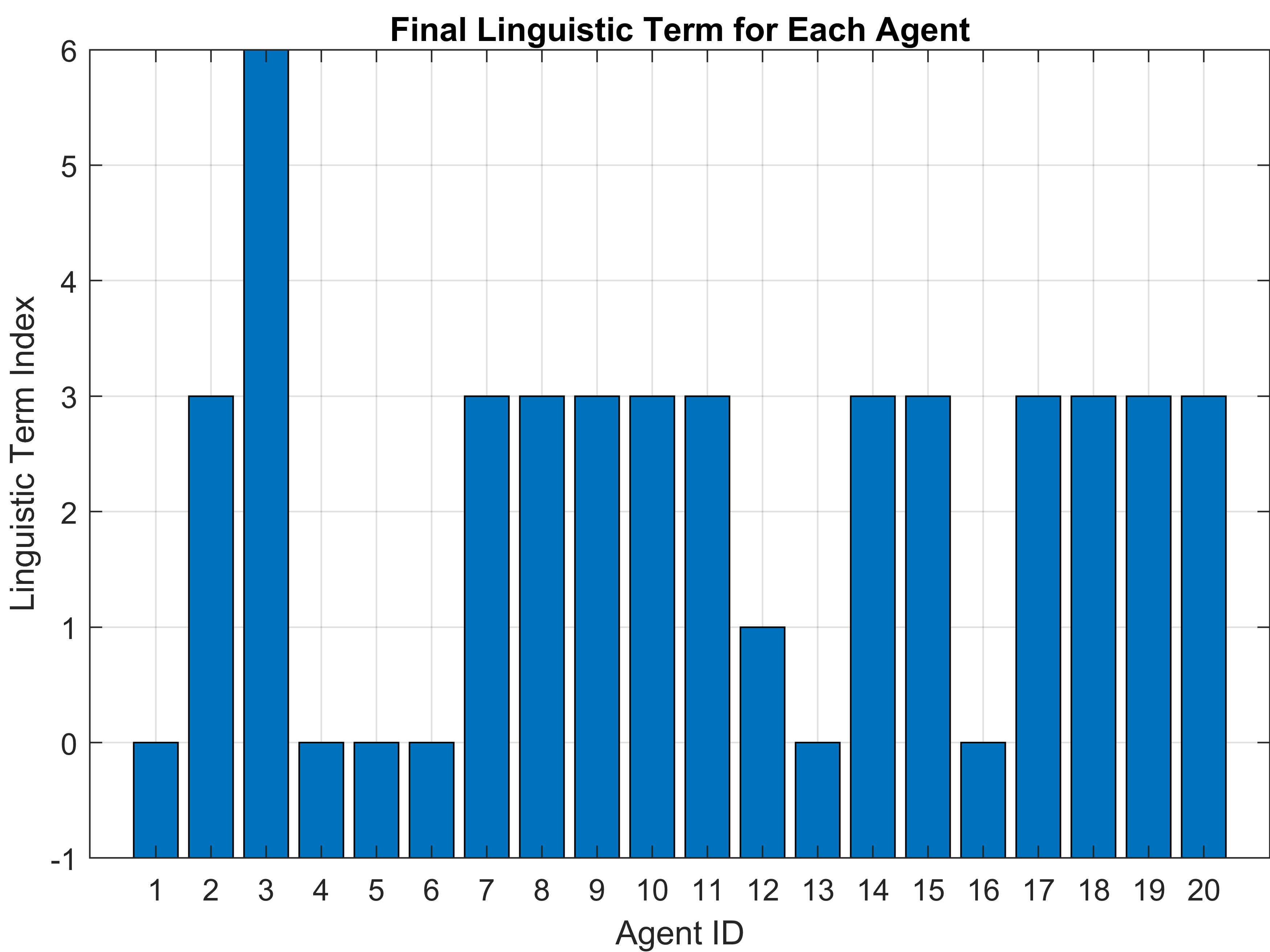}
\caption{Final Linguistic Term for Each Agent.}
\label{fig:final_resultsresults}
\end{figure}
\par \textcolor{black}{In the 3WD framework, where} $[\alpha, \beta]$ is the hesitation region, the choice of thresholds directly influences how likely agents are to exchange information. In Fig.~\ref{fig:final_resultsresults}, $\alpha$ is set to 0.2 and $\beta$ to 0.4, which defines a narrow acceptance range. Under this setting, only a small number of opinion differences fall within the acceptance or hesitation zones, while most are rejected directly. As a result, agents are more likely to avoid opinion exchange, leading to more isolated updates and a wider spread in final linguistic term indices. This \textcolor{black}{produces more diverse} and polarized results. In contrast, Fig.~\ref{fig:final_results} uses $\alpha = 0.3$ and $\beta = 0.6$, expanding both the acceptance and hesitation regions. More agent pairs are allowed to interact, either certainly or probabilistically, which leads to smoother opinion integration. Therefore, smaller $\alpha$ and $\beta$ reduce communication opportunities and promote diversity, while larger thresholds encourage interaction and support consensus formation.

\par
\begin{figure}[!htp]
\centering
\includegraphics[width=3.0in]{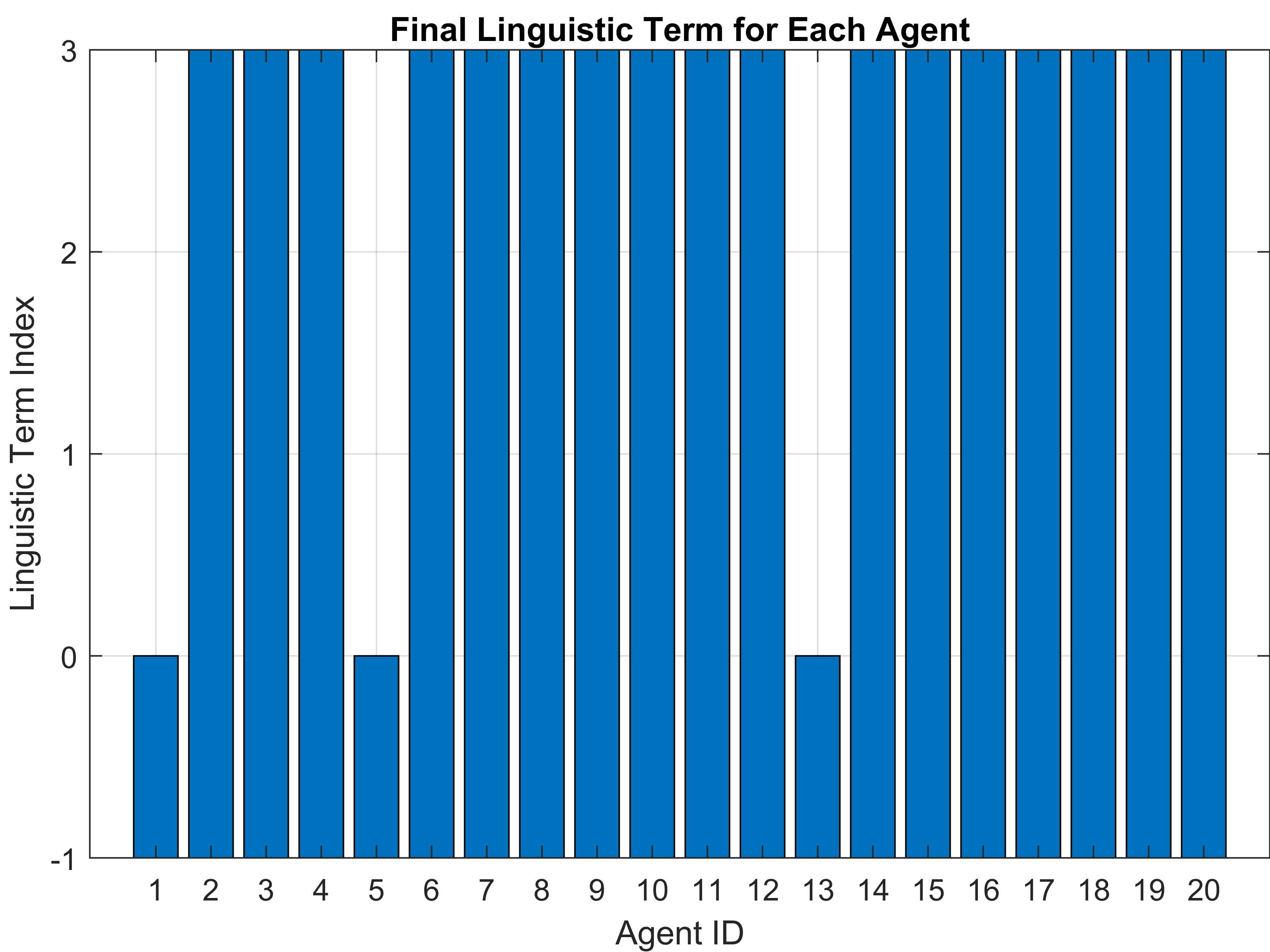}
\caption{Final Linguistic Term for Each Agent.}
\label{fig:final_resultsresultsresults}
\end{figure}

 In Fig.~\ref{fig:final_resultsresultsresults}, the thresholds are set to $\alpha = 0.5$ and $\beta = 0.9$, which significantly enlarges both the hesitation and acceptance regions compared to the previous settings. Under this configuration, only opinion differences greater than 0.9 are directly rejected, so most agent pairs have a chance to exchange information. As shown in the figure, the majority of agents reach a high and consistent linguistic term index $h_3$, while only a few remain at the lowest level $h_0$. \textcolor{black}{This result suggests that a wider acceptance rule enables more interaction among agents, facilitating their integration of opinions.} Compared to smaller values of $\alpha$ and $\beta$, this setup is more likely to support global agreement.

\subsubsection{Impact of Network Scale on Opinion Evolution and Consensus Formation}
\par
\begin{figure*}[!htp]
\centering
\includegraphics[width=7.0in]{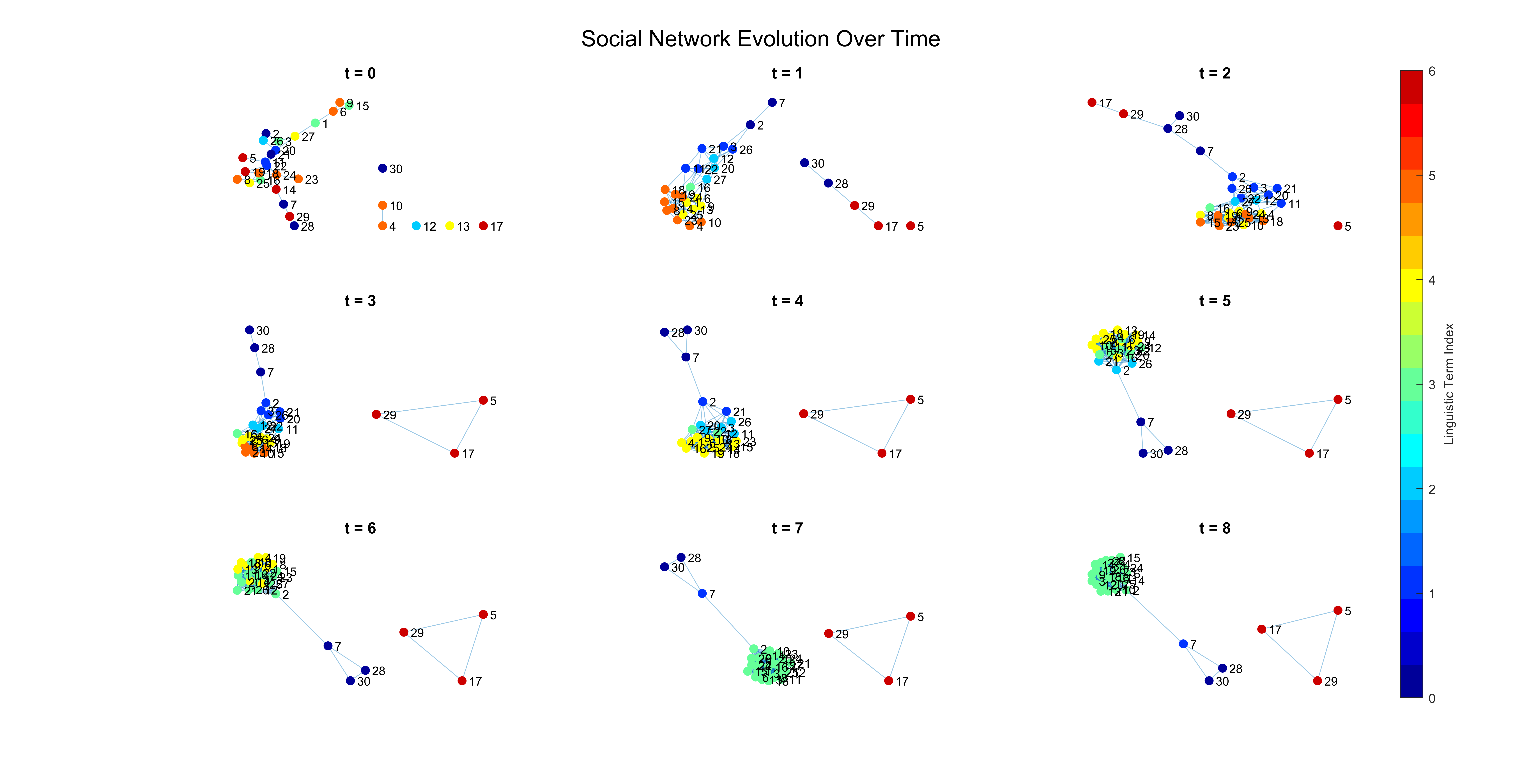}
\caption{Temporal Evolution of the Social Network and Linguistic Opinions.}
\label{fig:social_network1}
\end{figure*}

\begin{figure*}[!htp]
\centering
\includegraphics[width=7.0in]{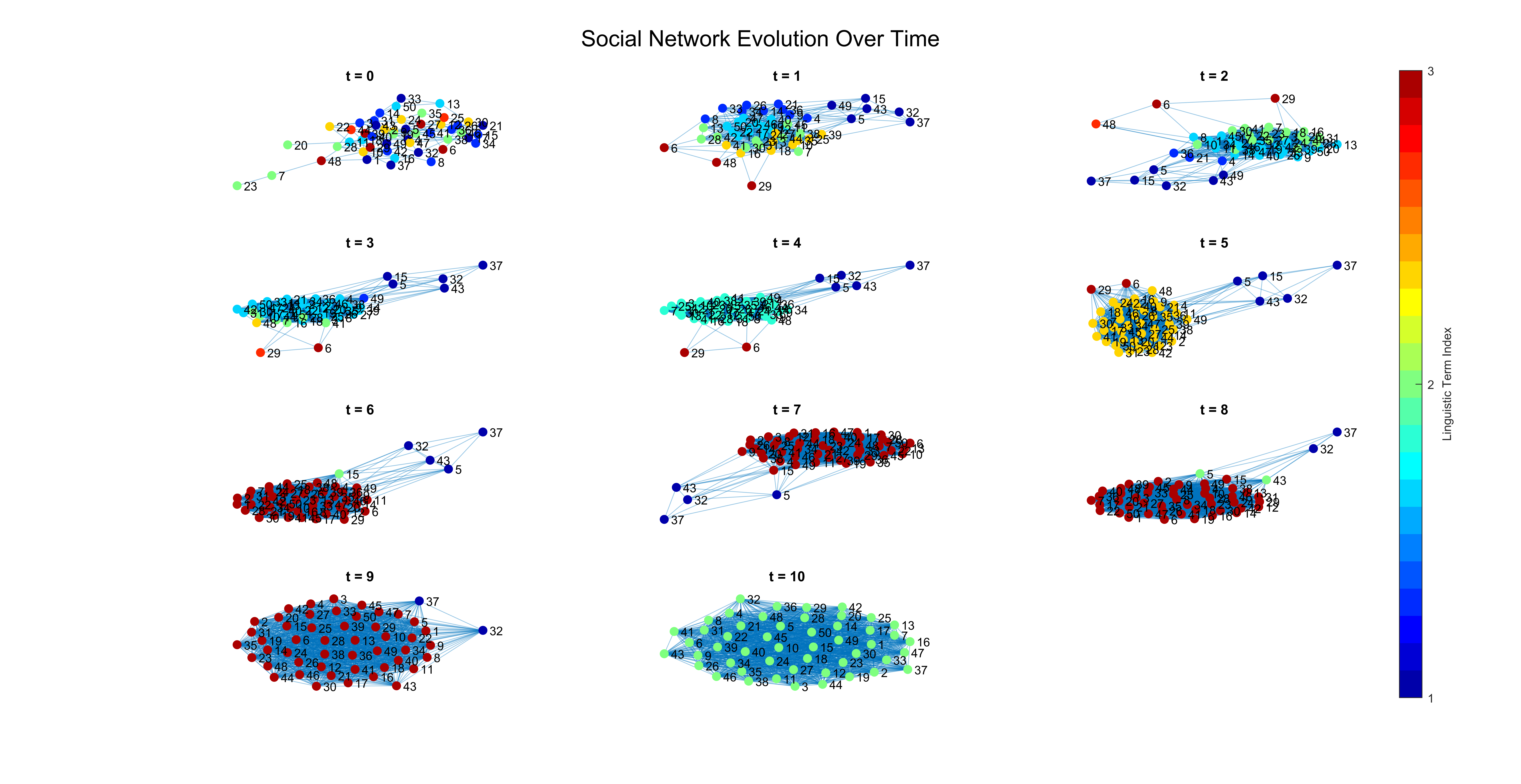}
\caption{Temporal Evolution of the Social Network and Linguistic Opinions.}
\label{fig:social_network2}
\end{figure*}

 Figs.~\ref{fig:social_network}, \ref{fig:social_network1}, and \ref{fig:social_network2} respectively illustrate the evolution process of the system when the number of agents is 20, 30, and 50, under the proposed model with 3WD mechanism and dynamic social network reconstruction rules. As the number of agents increases, the initially sparse social network gradually evolves into several closely connected subgroups. When $N=20$, the connections between agents are relatively concentrated, and two consensus groups are eventually formed. When $N=30$, the trend of network division becomes stronger, and some agents with extreme opinions form isolated small groups disconnected \textcolor{black}{from the leading network}. In the large-scale system with $N=50$, although the initial connections are \textcolor{black}{very sparse, the leading} group gradually absorbs most agents through the self-adjusting structure of the \textcolor{black}{network. Eventually, it forms} a unified group with closer connections and more consistent opinions. This phenomenon fully demonstrates the effectiveness and advantages of the proposed algorithm. As the scale of the social network increases, differences between agents grow, and the uncertainty of connections rises. Traditional static network models are likely to face problems such as opinion polarization, information isolation, and failure to reach consensus. In contrast, our model introduces a dynamic linking mechanism based on opinion distance and a ``stay undecided" option from the 3WD theory, which flexibly handles agents near the decision boundary and significantly improves the system’s ability to reach consensus under complex conditions. As shown in the figures, this mechanism not only helps to reduce the trend of division caused by sparse networks and opinion differences but also enhances information flow and coordinated evolution within the group.

\subsubsection{Impact of Three-Way Decision Mechanism on Opinion Dynamics}
\par
\begin{figure}[!htp]
\centering
\includegraphics[width=3.5in]{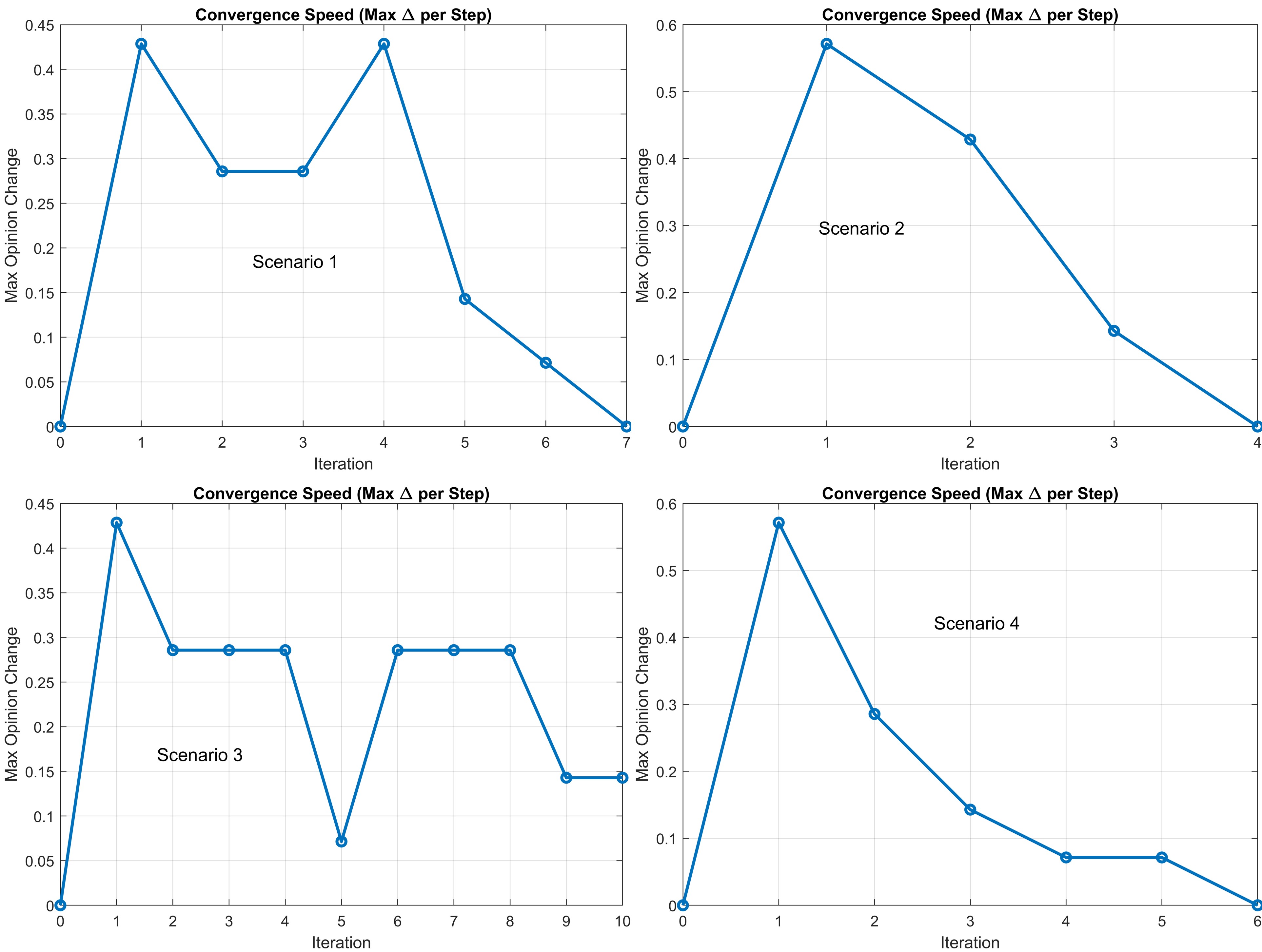}
\caption{Convergence Speed Comparison \\Under 
Different 3WD Settings.}
\label{fig:comparethreeway}
\end{figure}

 To evaluate the effect of the proposed 3WD mechanism on the evolution of group opinions, we conducted a comparative study involving four scenarios. Fig.~\ref{fig:comparethreeway} shows a comparison of convergence speed in four different scenarios. Each subplot represents a specific setting, where the horizontal axis is the number of iterations and the vertical axis shows the maximum change in opinions per step, denoted as $\delta_{\text{max}}$. This value is used as the measurement of convergence speed in all four experiments, and \textcolor{black}{the following formula calculates it}:
\[
\delta_{\text{max}}(t) = \max_i |\theta_i(t) - \theta_i(t-1)|
\]
Here, $\theta_i(t)$ represents the opinion of agent $i$ at $t$. This metric reflects the largest individual change at each step and helps evaluate how fast or smooth the opinion evolution process is. In the experiments, \textcolor{black}{Scenarios 1 and 3} represent cases with 40 and 60 agents, respectively, \textcolor{black}{where the 3WD mechanism} is applied using parameters $\alpha = 0.3$ and $\beta = 0.6$. Scenarios 2 and 4 use the same number of agents but do not include the 3WD mechanism, which is represented by $\alpha = \beta = 0.6$, meaning no hesitation zone is added. From the figure, we can see that in Scenarios 2 and 4 (without the mechanism), $\delta_{\text{max}}$ drops quickly, and convergence is reached within 4 and 6 iterations, respectively. The process is fast and smooth. In contrast, Scenarios 1 and 3 (with the mechanism) show a slower convergence process, with a more gradual decrease in $\delta_{\text{max}}$ and some fluctuations, especially in Scenario 3. Overall, the 3WD mechanism slows down the speed of opinion convergence, which results in a slower but more stable evolution. This effect becomes more visible when the number of agents increases. These results suggest that the 3WD mechanism achieves a balance between convergence speed and system stability.

\subsubsection{Impact of $\delta_{\text{add}}$ on Opinion Dynamics}
\par
\begin{figure*}[!htp]
\centering
\includegraphics[width=7.0in]{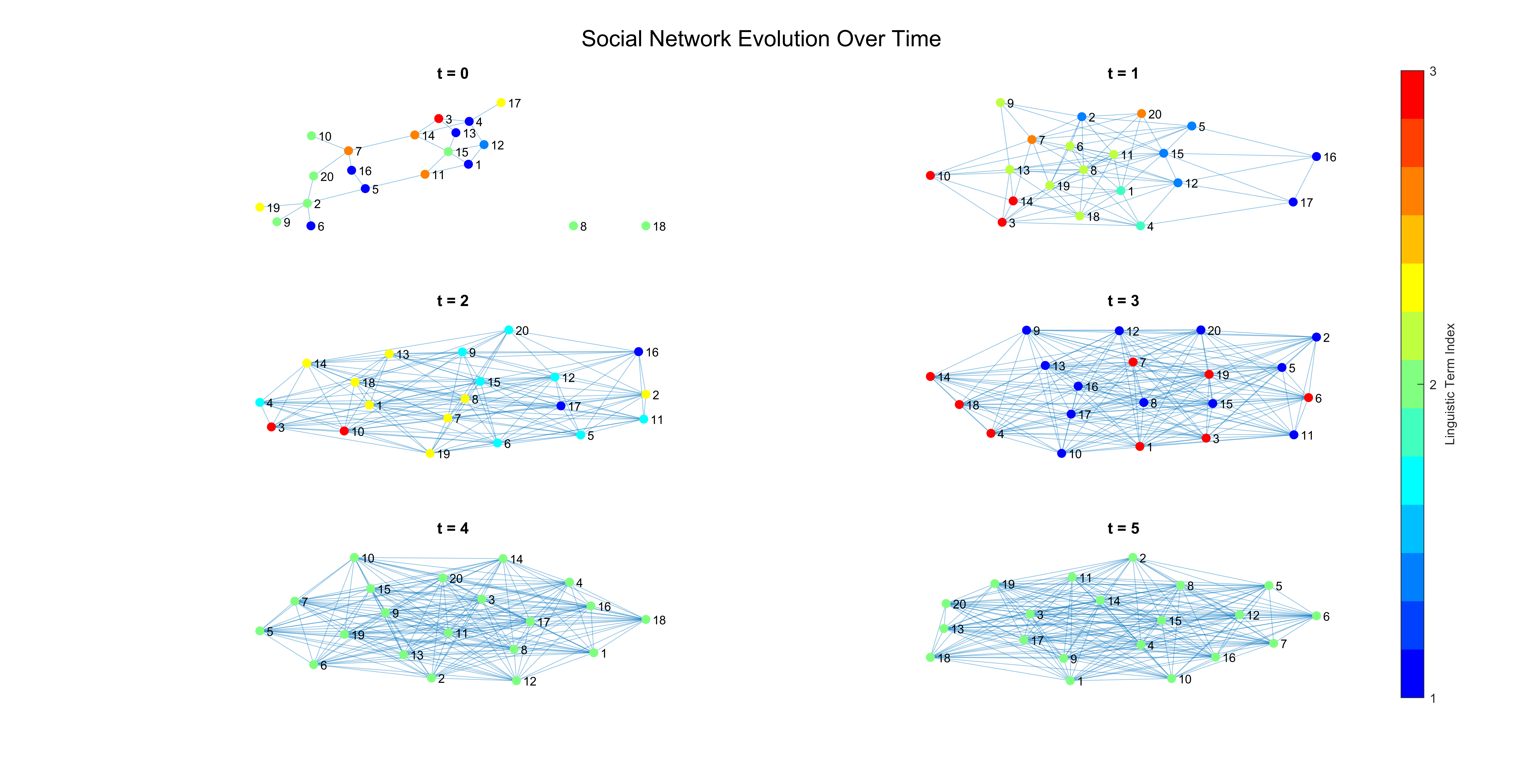}
\caption{Temporal Evolution of the Social Network and Linguistic Opinions.}
\label{fig:social_network_reconnect}
\end{figure*}

 Fig.~\ref{fig:social_network_reconnect} shows the evolution of the social network over time steps \(t = 0, 1, 2, 3, 4, 5\). The connection threshold \(\delta_{\text{add}}\) in Section Example is increased from 0.15 to 0.35. This higher value for adding links allows agents with larger opinion differences to form new connections, making it easier for the network to reconnect. From the figure, we observe that the initial network is fragmented with several isolated nodes. As time increases, agents rapidly form new connections and the network quickly becomes denser. By \(t = 3\), the system shows signs of global integration, and by \(t = 5\), all agents are part of a highly connected network with relatively homogeneous opinions. This result suggests that increasing \(\delta_{\text{add}}\) facilitates faster structural convergence, reduces isolation, and promotes opinion merging across the entire network.

\subsubsection{Impact of $\delta_{\text{cut}}$ on Opinion Dynamics}
\par
\begin{figure*}[!htp]
\centering
\includegraphics[width=7.0in]{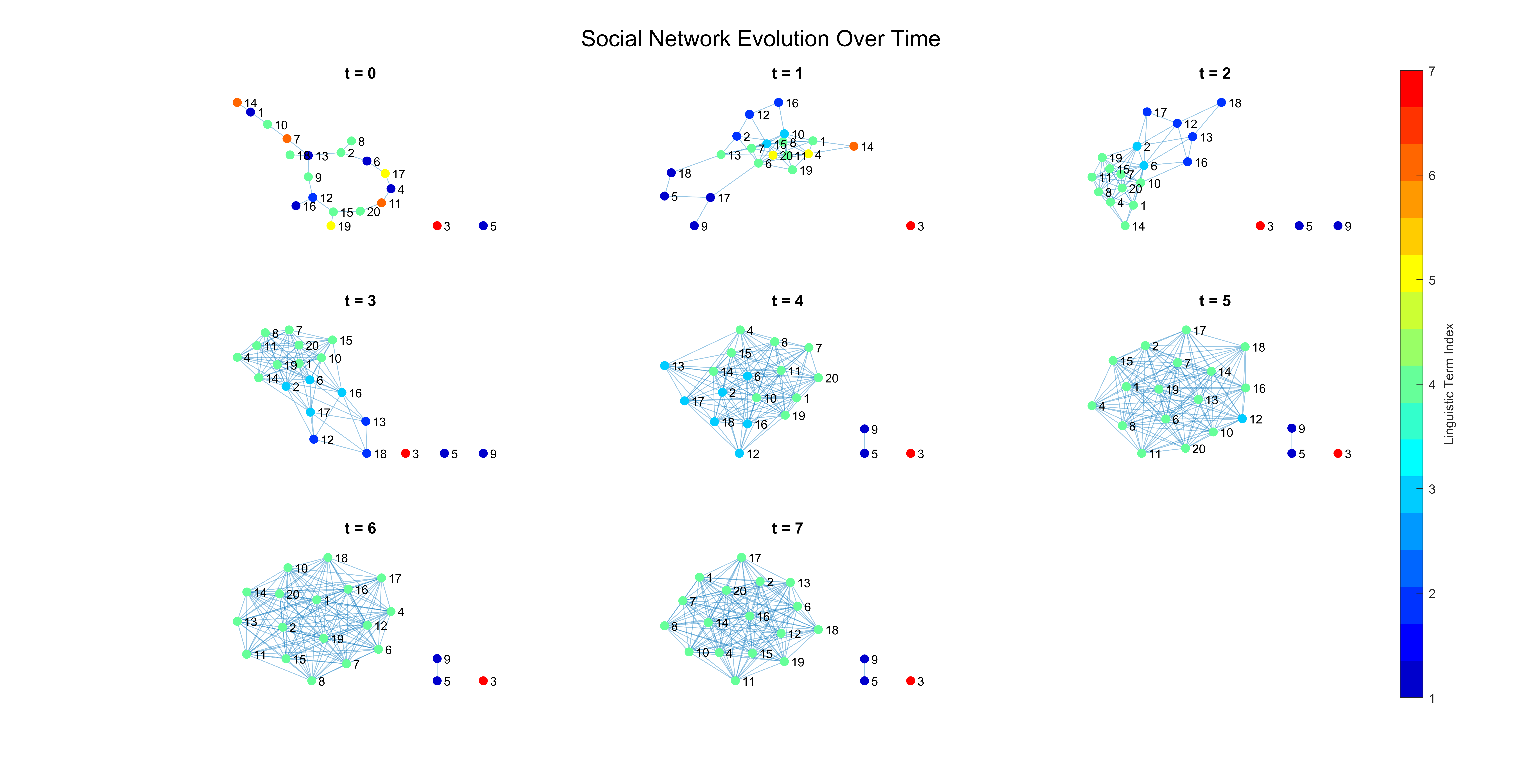}
\caption{Temporal Evolution of the Social Network and Linguistic Opinions.}
\label{fig:social_network_reconnectreconnect}
\end{figure*}

 Fig.~\ref{fig:social_network_reconnectreconnect} illustrates the evolution of the social network over time steps \(t = 0, 1, 2, \dots, 7\). In this example, we decreased the disconnection threshold \(\delta_{\text{cut}}\) from 0.45 to 0.2. This makes it easier for existing links to be removed when the opinion distance between agents exceeds the threshold, increasing the likelihood of network fragmentation. As shown in the figure, although the network starts as a connected structure, it quickly begins to break into separate components. By \(t = 2\), $e_3$, $e_5$, and $e_9$ become isolated from the main group. This pattern persists through later time steps, where the majority of agents form a dense cluster while the isolated ones remain disconnected. These results indicate that a lower \(\delta_{\text{cut}}\) can reduce network stability and reinforce persistent social separation.

\subsection{Complexity Analysis}
\subsubsection{Time Complexity} 
\par \textcolor{black}{In each iteration, the proposed model checks all possible neighbor pairs \((i, j)\) to compute opinion distances and determine acceptance. In the worst case, each agent needs to compare with \(N - 1\) others, resulting in a complexity of \(\mathcal{O}(N^2)\) for this step. The dynamic rewiring step also involves checking all agent pairs and may update the network structure, which again takes \(\mathcal{O}(N^2)\) time. The convergence check requires scanning all agents, which takes \(\mathcal{O}(N)\) time. Therefore, each iteration has a total time complexity of \(\mathcal{O}(N^2)\), and with a maximum of \(T_{\max}\) iterations, the overall time complexity is \(\mathcal{O}(T_{\max} \cdot N^2)\).}

\subsubsection{Space Complexity} 
\par \textcolor{black}{The primary source of memory usage is storing the network and agent states.} The adjacency matrix takes \(\mathcal{O}(N^2)\) space. Each agent’s numerical opinion \(\theta_i(t)\) and linguistic opinion \(x_i(t)\) together take \(\mathcal{O}(N)\) space. In addition, the neighbor sets may each contain up to \(N\) elements, which leads to a worst-case memory usage of \(\mathcal{O}(N^2)\). Therefore, the total space complexity is \(\mathcal{O}(N^2)\).
\par
\begin{figure}[!htb]
\centering
\includegraphics[width=3.5in]{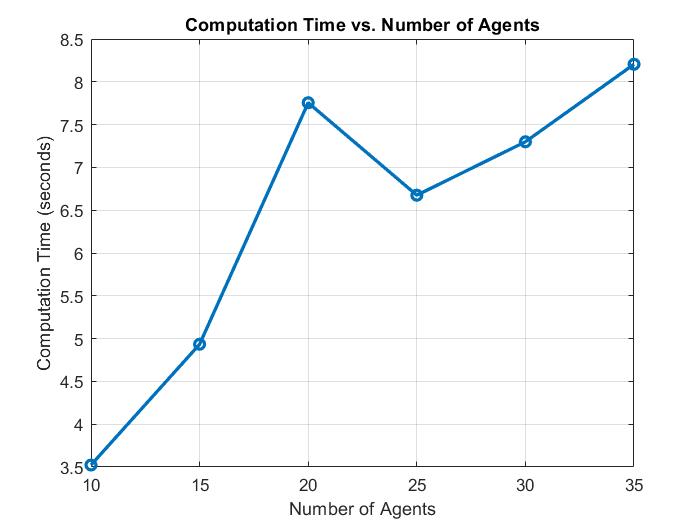}
\caption{Impact of Agent Number on Computation Time.}
\label{fig:computation_time}
\end{figure}

 \textcolor{black}{To test how computation time changes with the number of agents, we carried out experiments on the MATLAB 2020a platform, and the results are shown in Fig.~\ref{fig:computation_time}. Between 10 and 20 agents, the computation time increases clearly, from about 3.5 seconds to nearly 7.8 seconds. When the number of agents rises from 20 to 25, the time drops slightly to about 6.7 seconds. After that, between 25 and 35 agents, the time increases again and reaches about 8.2 seconds at 35 agents. \textcolor{black}{These results indicate that as the number of agents increases, the overall computation time grows, but the fluctuation is not significant and the growth is not strictly linear.} In summary, this algorithm models the information evolution with a reasonable level of computational complexity. Although the worst-case time and space complexity are both \(\mathcal{O}(N^2)\), real-world social networks are usually sparse and locally connected, so the actual runtime is often much lower than the theoretical upper bound. Moreover, the model has a clear structure and \textcolor{black}{is easily implementable} in parallel.}

\subsection{Comparative Analysis with Mainstream Models}
\par \textcolor{black}{In response to the reviewer’s suggestion, we conducted additional comparative experiments with several representative opinion dynamics models. The aim was to evaluate the proposed model against widely used approaches and to verify whether the observed behaviors are consistent or significantly different. All simulations in this section use the same initial opinions as those described in Section \textbf{Example}.}
\subsubsection{DeGroot Model}
\par \textcolor{black}{The DeGroot model is a classical linear consensus model in which each agent updates its opinion by taking a weighted average of all agents’ opinions, as introduced in \emph{Definition 3}. Here, two variants of the DeGroot model are considered:}
\begin{itemize}
    \item \textcolor{black}{Uniform weights: In this case, each agent assigns the same weight to every other agent, regardless of the opinion distance. This represents the most basic form of the DeGroot update rule.} 
    \item \textcolor{black}{Distance-based weights: In this variant, the interaction weight between two agents depends on the distance between their opinions. The weight matrix is calculated using $\omega_j = \frac{e^{-\vert D_{ij}\vert}}{\sum_{k=1}^{20} e^{-\vert D_{ik}\vert}}$. This formulation ensures that agents with more similar opinions exert a stronger influence on each other.}
\end{itemize}
\par
\begin{figure}[!htb]
\centering
\includegraphics[width=3.0in]{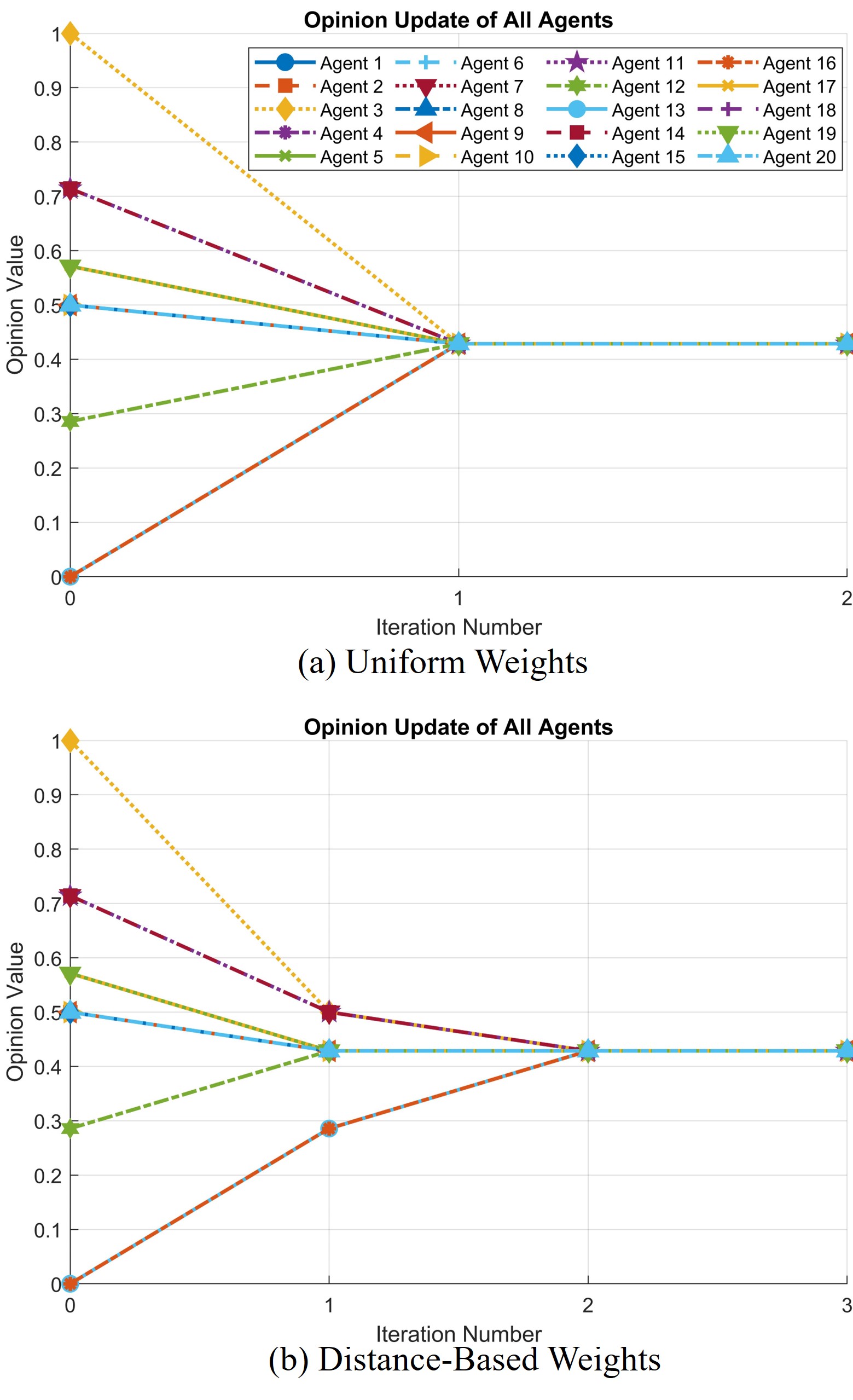}
\caption{Opinion Evolution under the DeGroot Model with Uniform and Distance-Based Weights.}
\label{fig:DG}
\end{figure}

 \textcolor{black}{Under the DeGroot model, we compare two methods: uniform weights and distance-based weights. The results are shown in Fig.~\ref{fig:DG}. In the uniform weight case, the weight of each agent \textcolor{black}{is the same, 1/20}. Since each row of the weight matrix is identical, the system quickly converges after the first iteration to the arithmetic mean of all initial \textcolor{black}{opinions. It remains} unchanged, showing an averaging effect and one-step convergence. In the distance-based weight case, agents with higher opinion similarity have a stronger influence on each other. In the early stage, agents with extreme opinion values change faster, while those in the middle change more slowly. The system becomes stable after 2–3 iterations, and the final consensus value is the weighted average of all initial opinions, which is slightly different from the simple average in the uniform weight case. These results show that the DeGroot model is sensitive to the way the weights are calculated.}

\subsubsection{Homogeneous HK Bounded Confidence Model}

\begin{figure*}[!htb]
\centering
\includegraphics[width=7.0in]{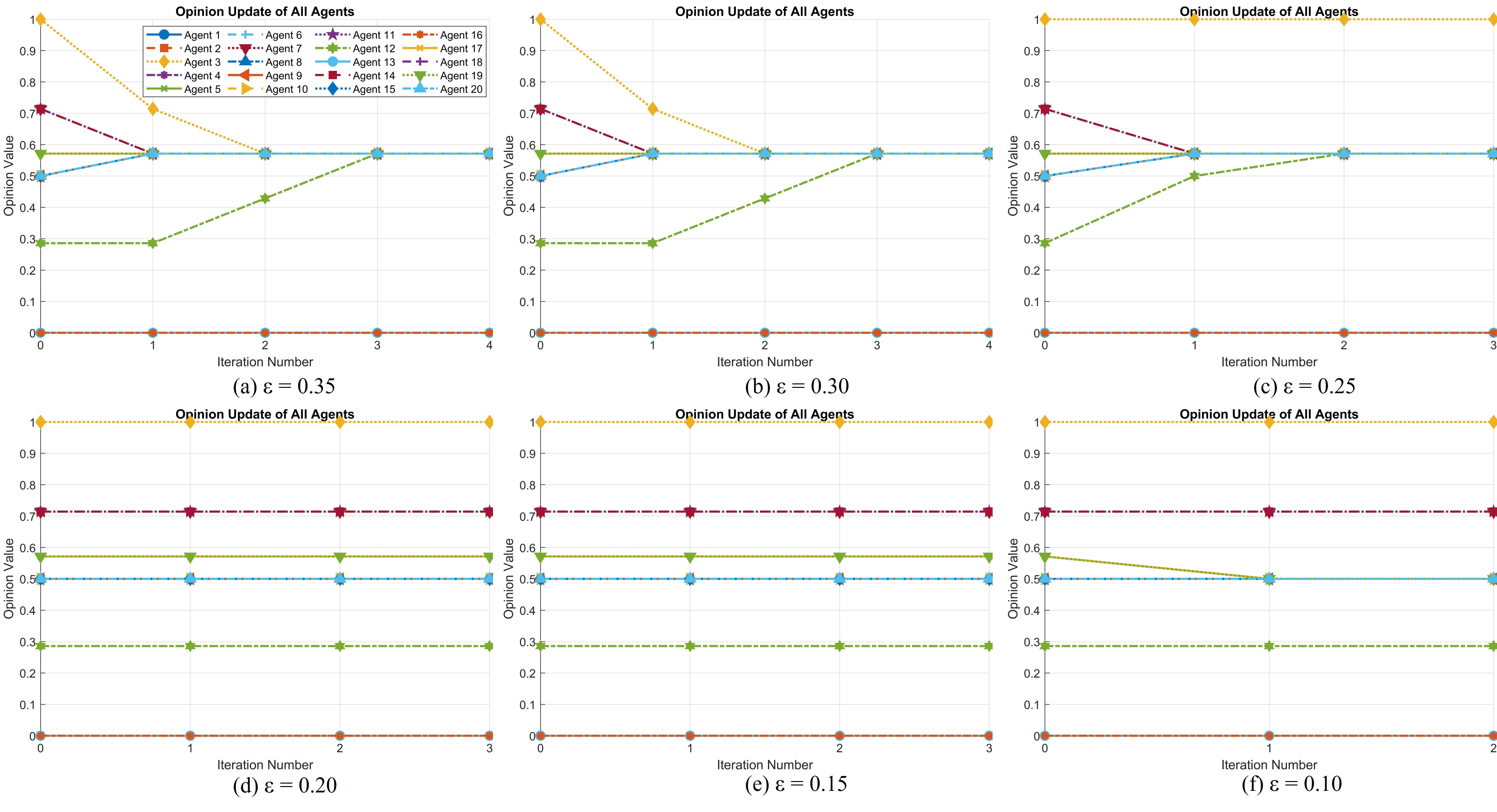}
\caption{Opinion Evolution of All Agents under Different Confidence Bounds in the Homogeneous HK Model.}
\label{fig:HK_same}
\end{figure*}

The homogeneous HK bounded confidence model assumes that all agents share the same confidence bound \(\epsilon\), as introduced in \emph{Definition 4}. At each iteration, an agent only considers the opinions of others whose opinion difference is less than \(\epsilon\). To enable a thorough comparative analysis, we tested six different values of \(\epsilon\), namely \(0.35\), \(0.30\), \(0.25\), \(0.20\), \(0.15\), and \(0.10\). As \(\epsilon\) decreases, the communication threshold between agents becomes smaller, meaning that an agent interacts only with those whose opinions are very similar to its own. Fig.~\ref{fig:HK_same} illustrates the opinion evolution of all agents under six different confidence bound values in the homogeneous HK bounded confidence model. When \(\epsilon = 0.35\) [Fig.~\ref{fig:HK_same}(a)], several agents are within each other’s confidence intervals, leading to active opinion exchange across multiple groups. This interaction results in partial convergence toward intermediate values after a few iterations. When \(\epsilon = 0.30\) [Fig.~\ref{fig:HK_same}(b)], the interaction range slightly decreases. However, since most of the key interaction links remain, the evolution results are the same as in the \(\epsilon = 0.35\) case. For \(\epsilon = 0.25\) [Fig.~\ref{fig:HK_same}(c)], the threshold is further reduced, causing only a subset of agents to interact. Opinion exchange occurs primarily within smaller local clusters, and the number of isolated agents increases. When \(\epsilon = 0.20\) [Fig.~\ref{fig:HK_same}(d)], \(\epsilon = 0.15\) [Fig.~\ref{fig:HK_same}(e)], and \(\epsilon = 0.10\) [Fig.~\ref{fig:HK_same}(f)], as the threshold gradually decreases, the information exchange between agents becomes less frequent, making it difficult for them to reach consensus in the end. \textcolor{black}{Through the analysis, it is evident that this model is susceptible to the threshold, where even slight differences can result in significant variations in the results.}

\subsubsection{Heterogeneous HK Bounded Confidence Model}

\textcolor{black}{The heterogeneous HK bounded confidence model extends the homogeneous version by assigning different threshold values to agents. In the first case, the threshold $\epsilon_i$ of each agent is listed below:}
\textcolor{black}{\[
\begin{aligned}
\epsilon_1 &= 0.2, \quad &\epsilon_2 &= 0.5, \quad &\epsilon_3 &= 0.3, \quad &\epsilon_4 &= 0.4, \\
\epsilon_5 &= 0.2, \quad &\epsilon_6 &= 0.1, \quad &\epsilon_7 &= 0.9, \quad &\epsilon_8 &= 0.6, \\
\epsilon_9 &= 0.5, \quad &\epsilon_{10} &= 0.3, \quad &\epsilon_{11} &= 0.2, \quad &\epsilon_{12} &= 0.1, \\
\epsilon_{13} &= 0.4, \quad &\epsilon_{14} &= 0.4, \quad &\epsilon_{15} &= 0.5, \quad &\epsilon_{16} &= 0.3, \\
\epsilon_{17} &= 0.7, \quad &\epsilon_{18} &= 0.4, \quad &\epsilon_{19} &= 0.2, \quad &\epsilon_{20} &= 0.2.
\end{aligned}
\]}
\par
\begin{figure*}[!htb]
\centering
\includegraphics[width=7.0in]{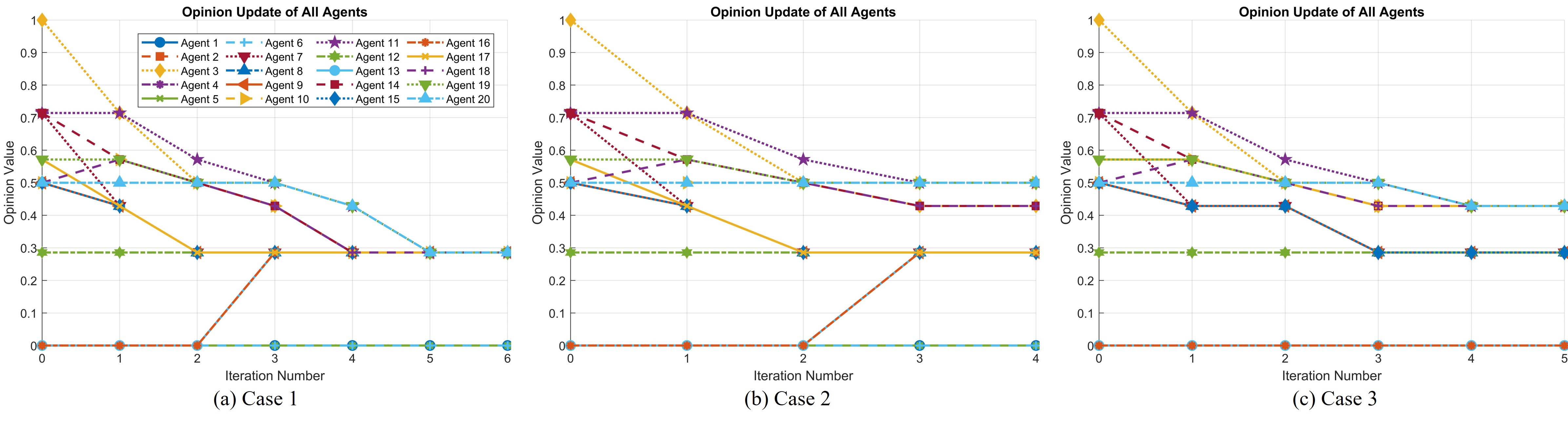}
\caption{Opinion Evolution of All Agents under Different Confidence Bounds in the Heterogeneous HK Model.}
\label{fig:HK_different}
\end{figure*}

 \textcolor{black}{In the second case, $\epsilon_{10}$ is changed from 0.3 to 0.2, while all other values remain unchanged. In the third case, $\epsilon_{17}$ is changed from 0.7 to 0.3, while all other values remain unchanged. The results are shown in Fig.~\ref{fig:HK_different}. \textcolor{black}{The final results are distinctly different}. Case 1 converges to two distinct opinion clusters, Case 2 produces a different set of stable clusters, and Case 3 results in yet another opinion distribution. These results demonstrate that the model is also susceptible to individual parameter values, where even a small change can lead to significantly different results.}

\subsection{Potential Applications}
\subsubsection{Brain-Computer Interface (BCI) systems} 
\par BCI systems often involve non-invasive brain signals such as electroencephalography (EEG) and functional near-infrared spectroscopy (fNIRS). \textcolor{black}{These signals typically exhibit high variability, significant individual differences, and incomplete information}. Traditional classification methods require making clear decisions for all states. However, in real-world applications, the user's cognitive state may be ``uncertain" or ``hesitant." Introducing the 3WD theory allows the brain signal recognition system to produce three types of outputs: “accept,” “reject,” and “defer.” This keeps an intermediate state for handling vague information and helps avoid incorrect triggering. In addition, the mechanism of dynamically adjusting neural information channels can also inspire the design of adaptive models for individual brain region communication.
\begin{figure}[!htb]
\centering
\includegraphics[scale=0.7]{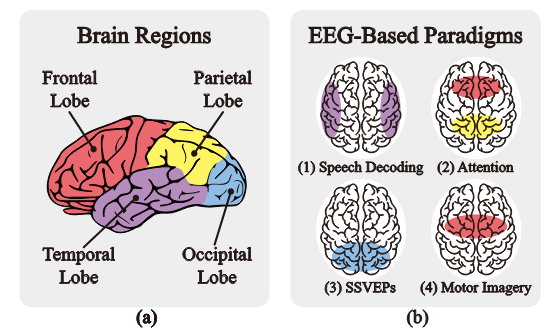}
\caption{Description of Brain Regions and EEG-Based
Paradigms. (a) Four Regions of the Cortex. (b) Four Other Common EEG-Based Paradigms \cite{luo2019upefedatfeing}.}
\label{debris}
\end{figure}

\subsubsection{Opinion Fusion in Smart City Governance} 
\par \textcolor{black}{In smart city governance}, such as public policy making, intelligent traffic management, and urban risk warning, a large amount of public opinion needs to be collected and integrated. When there are differences in views, uncertain perceptions, and vague opinions, traditional opinion fusion models can easily lead to extreme results or distorted information. The proposed model allows individual opinions to be expressed in natural language and handles uncertain areas through deferred decisions. \textcolor{black}{This makes the opinion fusion process more comprehensible and less prone to ambiguity.} 

\subsubsection{Multi-Robot Coordination in Disaster Response and Agricultural Monitoring}
\par In scenarios such as post-disaster search and rescue and agricultural monitoring, multi-robot systems \textcolor{black}{must make} collaborative decisions based on environmental sensing, \textcolor{black}{their own capabilities}, and task requirements. Traditional assignment strategies often require each robot \textcolor{black}{to accept or reject a task immediately}. However, in uncertain environments or when task complexity is unclear, this approach may lead to poor task allocation or wasted resources. By introducing the 3WD theory, robots are allowed to delay task decisions, so they can make better choices when more information becomes available. \textcolor{black}{At the same time, by dynamically adjusting the communication network, robots can enhance their cooperation based on task similarity or path overlap, thereby increasing the overall efficiency and robustness of the system.}

\section{Conclusion}
\par This paper proposes a group opinion update model that combines the 3WD mechanism with a dynamic network update process. The introduced 3WD mechanism sets a ``hesitation interval" to expand the traditional binary rule of accepting or rejecting neighbors into a three-way probabilistic rule. This allows an agent to accept information from another agent with a certain probability when their opinion difference lies in an uncertain range. In addition, to overcome the limitations of a static network structure, the model includes a dynamic update mechanism based on cognitive similarity. By setting probabilistic rules for adding or removing links, the social relationships between agents can adjust in real time as their opinions evolve. The proposed algorithm is applied to a multi-UAV cooperative decision-making scenario. By simulating the information exchange and dynamic negotiation among UAVs, we demonstrate that the mechanism is effective in practical distributed intelligent systems. The experimental comparisons further verify the advantages of the algorithm.

\par As AI agents powered by large language models (LLMs) play an increasingly important role in group decision-making, how to combine the diverse solutions generated by different models has become a key issue. Future research can focus on conflict resolution between models, trust evaluation mechanisms, and integration strategies to improve the overall level of collaborative intelligence in the system. \textcolor{black}{Additionally, we will incorporate reinforcement learning or adaptive weighting based on past interactions to enable agents to refine their decision strategies over time}.

\bibliography{mybibfile}

\end{document}